\documentclass[runningheads]{llncs}
\usepackage[T1]{fontenc}
\usepackage{graphicx}
\usepackage{booktabs}
\usepackage[misc]{ifsym}
\newcommand{\corr}{(\Letter)}
\newcommand{\repeatthanks}{\textsuperscript{\thefootnote}}
\usepackage{mwe}
\usepackage{subcaption}
\usepackage{tabularx}

\newcolumntype{C}{>{\centering\arraybackslash}X}
\usepackage{multirow}
\usepackage{rotating}
\usepackage{float}
\usepackage{longtable,verbatim}
\usepackage{xcolor}
\usepackage{amsmath}
\usepackage{listings}
\usepackage{tcolorbox}
\usepackage{hyperref}

\begin{document}

\title{Beyond the Visible: Multispectral Vision-Language Learning for Earth Observation}

\titlerunning{Multispectral Vision-Language Learning for Earth Observation}

\author{Clive Tinashe Marimo \thanks{Equal contribution.}\inst{1} \and
Benedikt Blumenstiel \repeatthanks\inst{2} \corr \and
Maximilian Nitsche\inst{1} \and
Johannes Jakubik\inst{2} \and
Thomas Brunschwiler\inst{2}}
\authorrunning{Marimo, C., Blumenstiel, B., Nitsche, M., Jakubik, J., and Brunschwiler, T.}

\institute{
IBM Germany \and
IBM Research Europe \\\email{benedikt.blumenstiel@ibm.com}}

\tocauthor{Clive Tinashe Marimo, Benedikt Blumenstiel, Maximilian Nitsche, Johannes Jakubik, Thomas Brunschwiler}
\toctitle{Beyond the Visible: Multispectral Vision-Language Learning for Earth Observation}

\maketitle

\begin{abstract}
Vision-language models for Earth observation (EO) typically rely on the visual spectrum of data as the only model input, thus failing to leverage the rich spectral information available in the multispectral channels recorded by satellites. Therefore, we introduce Llama3-MS-CLIP––the first vision-language model pre-trained with contrastive learning on a large-scale multispectral dataset and report on the performance gains due to the extended spectral range. Furthermore, we present the largest-to-date image-caption dataset for multispectral data, consisting of one million Sentinel-2 samples and corresponding textual descriptions generated using Llama3-LLaVA-Next and Overture Maps data. We develop a scalable captioning pipeline, which is validated by domain experts. We evaluate Llama3-MS-CLIP on multispectral zero-shot image classification and retrieval using three datasets of varying complexity. Our results demonstrate that Llama3-MS-CLIP significantly outperforms other RGB-based approaches, improving classification accuracy by +6.77\% on average and retrieval performance by +4.63\% mAP compared to the second-best model. Our results emphasize the relevance of multispectral vision-language learning. The image-caption dataset, code, and model weights are available at \url{https://github.com/IBM/MS-CLIP}.

\end{abstract}

\section{Introduction}

Vision-language models (VLM) have transformed computer vision, enabling powerful zero-shot learning and cross-modal retrieval capabilities~\cite{openaiclip,openclip}. By learning joint representations of images and text, these models generalize across tasks without requiring task-specific training data.However, existing VLMs, such as CLIP~\cite{openaiclip}, are predominantly trained on natural RGB images, limiting their applicability to specialized domains such as Earth observation(EO)~\cite{georsclip,skyclip,remoteclip}. Conversely, effectively utilizing multispectral input data in VLMs represents an interesting and underexplored research topic in the machine learning community.

\begin{figure}[tbh]
    \centering
    \includegraphics[width=0.8\textwidth]{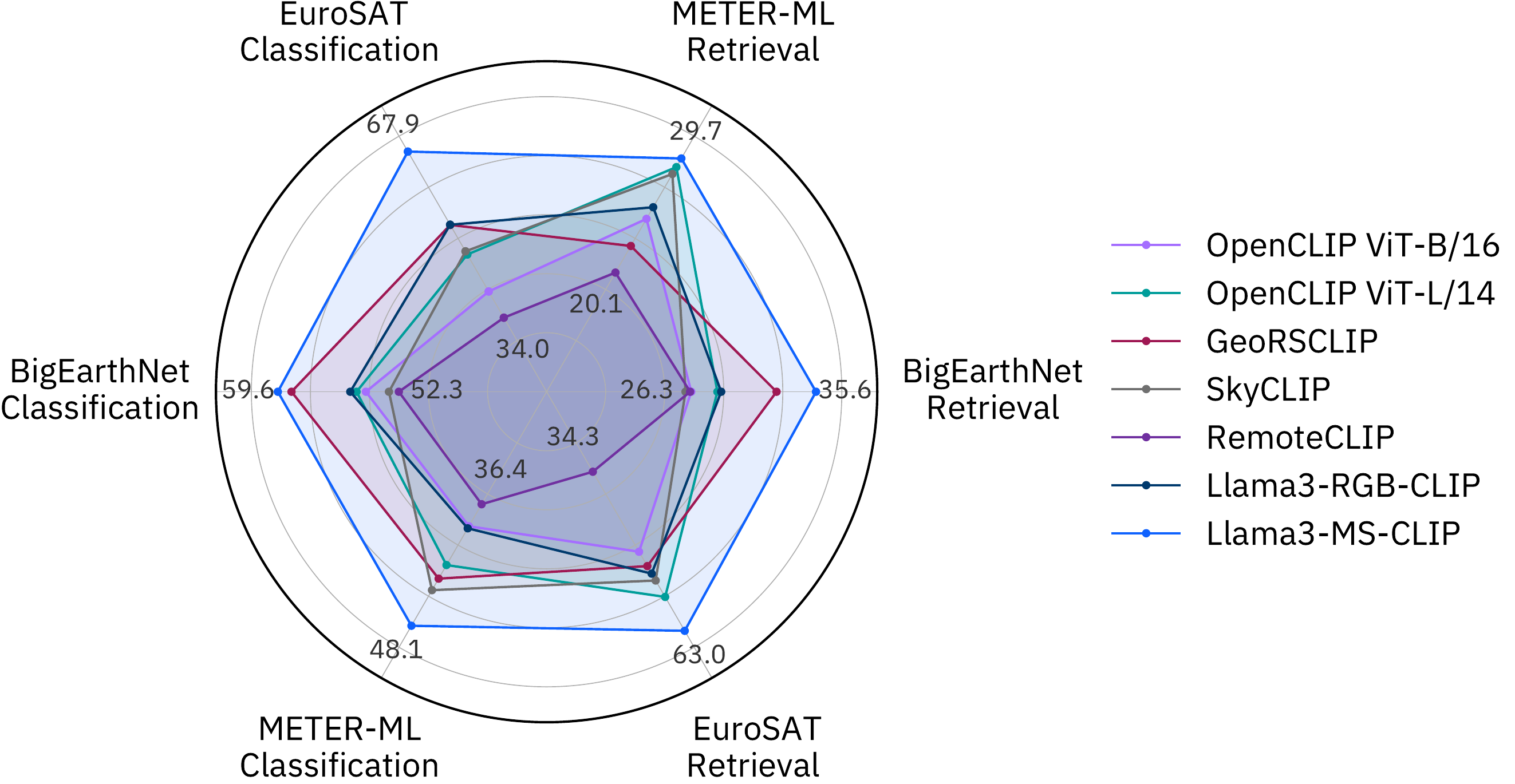}
    \caption{Zero-shot classification and text-to-image retrieval results, measured in accuracy (\%) $\uparrow$ and mAP@100 (\%) $\uparrow$, respectively. We applied a smoothed min-max scaling and annotated the lowest and highest scores. The multispectral CLIP is outperforming other RGB-based models on most benchmarks.}
    \label{fig:radar}
\end{figure}

EO relies on satellite imagery to monitor environmental changes, urban expansion, and agriculture~\cite{zhu2017deep}. To use VLMs for such EO applications, researchers proposed a range of domain-specific EO adaptations of CLIP~\cite{georsclip,skyclip,remoteclip,li2023rs}. Those models have been trained on up to five million remote sensing images with aligned image captions. Despite impressive performance, these models rely only on RGB input channels instead of leveraging the full spectral range available in multispectral (MS) satellite data for maximum effectiveness. 
Satellites like Sentinel-2 provide up to 13 spectral bands, capturing rich information far beyond visible wavelengths. Yet, until today, no large-scale multispectral dataset with image captions is publicly available for the research community.

To address this gap, we generate approximately one million captions for a popular, multispectral EO dataset derived from the Sentinel satellites called SSL4EO-S12~\cite{ssl4eos12}. We develop an automated captioning approach using metadata tags from Overture maps and the multimodal large language model (MLLM) Llama3-LLaVA-Next-8B~\cite{llavanext}. These captions provide semantic grounding for contrastive learning, allowing the model to align multispectral image representations with natural language. We then perform continual pre-training on OpenCLIP~\cite{openclip}, adapting the model to the EO domain.
Our model, Llama3-MS-CLIP\footnote{Built with Meta Llama 3. While the model itself is not based on Llama 3 but OpenCLIP B/16, it is trained on captions generated by a Llama 3-derivative model. Therefore, the model name starts with Llama 3 following its license (\url{https://github.com/meta-llama/llama3/blob/main/LICENSE}).}, outperforms other VLMs on a range of downstream applications as depicted in Figure~\ref{fig:radar}. The results demonstrate that multispectral data significantly enhances vision-language learning in EO, unlocking capabilities that RGB-based models fail to capture.

Our contributions are threefold: (1) We create the largest multispectral image-caption dataset for EO, (2) we present the first multispectral EO VLM, surpassing current state-of-the-art performance, and (3) we propose novel best practices for model development and image  
captioning with multispectral data. The dataset, code, and model weights are available at \url{https://github.com/IBM/MS-CLIP} under a permissive license.

\section{Related Work}

Vision-language models have successfully enabled zero-shot learning and cross-modal retrieval by aligning image and text embeddings through contrastive learning. CLIP~\cite{openaiclip}, OpenCLIP~\cite{openclip}, and ALIGN~\cite{align} are among the most prominent models in this space, trained on vast datasets of internet-scale image-text pairs. These models excel at general vision tasks but are inherently biased toward natural RGB images. Their ability to generalize to remote sensing imagery is limited due to the domain gap between natural images and satellite imagery~\cite{openaiclip,georsclip}. Spectral information beyond the visible range is key for applications like vegetation and disaster monitoring, as well as urban planning, which benefit from near-infrared and short-wave infrared reflectance measurements~\cite{prithvi2}.

One of the major challenges in EO vision-language learning is the lack of large-scale image-text datasets with multispectral data~\cite{chatearthnet}. Unlike natural images, satellite images do not inherently come with descriptive text. Most EO datasets provide only categorical labels or metadata, making it challenging to train models that require diverse textual supervision. While some approaches have attempted to generate captions for EO images, they often rely on metadata-based descriptions~\cite{skyclip} or manually curated annotations~\cite{remoteclip}. These approaches do not scale to the large volumes of data required to train robust vision-language models.
UCMC~\cite{ucmc}, RSICD~\cite{rsicd}, and RSITMD~\cite{rsitmd} are some famous examples of human-curated EO datasets that are often limited by their small size.
RS5M~\cite{georsclip} collected five million images from eleven source datasets and used BLIP-2~\cite{blip2} to generate captions for the RGB images. RSCLIP~\cite{li2023rs} introduced a pseudo-labeling technique that automatically generates pseudo-labels from unlabeled data. ChatEarthNet~\cite{chatearthnet} is the first multispectral dataset with over 100k samples, using ChatGPT-3.5 to generate captions. However, the captions are solely based on a small set of land-cover classes without visual input for the LLM.

Recent approaches have adapted VLMs to EO by continual pre-training on domain-specific RGB datasets. For instance, SkyCLIP~\cite{skyclip}, RSCLIP~\cite{li2023rs}, and RemoteCLIP~\cite{remoteclip} built large-scale image-text datasets and adapted CLIP-based backbones. GeoRSCLIP~\cite{georsclip} was pre-trained on RS5M, the largest known image-test dataset in the domain with five million images. GRAFT~\cite{mall2023remote} utilized co-located street-view images to correlate satellite imagery with language. Despite these advancements, all these methods rely solely on RGB data, ignoring the rich spectral information available in multispectral EO imagery.We further note the emergence of autoregressive approaches that generate language output based on optical or radar images, like in TerraMind \cite{terramind}.

In contrast to prior work, we introduce a self-supervised approach based on multimodal large language models (MLLM) and Overture annotations to automatically generate captions for multispectral EO imagery (i.e., Sentinel-2 images). By fine-tuning OpenCLIP on this dataset, we enable multispectral vision-language learning, allowing the model to leverage spectral information beyond the visible spectrum. 

\section{Automated Captioning}

The effectiveness of vision-language models relies heavily on the availability of high-quality image-text datasets. Thus, we introduce Llama3-SSL4EO-S12 captions\footnote{Captions are available at \url{https://huggingface.co/datasets/ibm-esa-geospatial/Llama3-SSL4EO-S12-v1.1-captions}.}, a novel dataset of text data aligned with SSL4EO-S12\,v1.1~\cite{ssl4eos12_v11}. It provides detailed natural language descriptions required for contrastive learning of multispectral vision-language models. 

SSL4EO-S12\,v1.1~\cite{ssl4eos12_v11} consists of 975k co-registered images of optical data from Sentinel-2~L1C (top-of-atmosphere) and Sentinel-2~L2A (bottom-of-atmosphere) as well as synthetic aperture radar (SAR) data from Sentinel-1~GRD. The dataset covers 244k global locations centered around urban areas, with a 264~$\times$~264 pixel size at 10\,m resolution, each including samples from four seasons. 

We generate captions by employing a multimodal large language model, specifically Llama3-LLaVA-Next-8B. The model was selected based on a qualitative comparison and a quantitative evaluation of three MLLM models using METEOR (Metric for Evaluation of Translation with Explicit Ordering)~\cite{meteor}, comparing ground truth captions with generations. We tested BLIP2~\cite{blip2}, used for the captioning in the RS5M~\cite{georsclip} dataset, Llama3-LLaVA-Next-8B~\cite{llavanext}, and RS-LLaVA~\cite{rsllava}, a domain-specific adaptation of LLaVA 1.5. We assessed these models using UCM Captions~\cite{ucmc}, RSICD~\cite{rsicd}, and RSITMD~\cite{rsitmd} that provide human-annotated captions. Llama3-LLaVA-Next-8B reaches an average METEOR score of 0.20 compared to 0.16 for RS-LLaVA and 0.10 for BLIP2. All scores and some examples are provided in the supplementary material.

The captioning process consists of the following steps:
First, we extract the RGB channels from S-2~L2A data and scale it to a uint8 value range of 0––255 as no publicly available MLLM supports multispectral inputs. The images are resized to 224~$\times$~224 pixels as input for the captioning model.
We then extract geographical tags from the Overture Maps base layer\footnote{Overture Maps: \url{https://docs.overturemaps.org} (Version: 2024-03-12-alpha.0)} that provides additional contextual information about land cover, infrastructure, and other features in the satellite image. The geographical instances are then sorted and filtered by size, i.e., all features smaller than 2500 square meters (5~$\times$~5 pixels) are omitted. We use all tags of each instance as they include additional information, like intermittent rivers. We further add the names of places to avoid hallucinations. Otherwise, we observe that the model often refers to popular places incorrectly, like labeling most universities as \textit{Harvard} or \textit{Berkeley}.
Finally, we prompt the MLLM to generate captions in a structured manner by following a chain-of-thought approach. First, the model is prompted to generate three relevant question-answer pairs, guiding the model in producing the final caption. The prompt includes further instructions to avoid hallucinations and increase the caption quality. We repeat the generation if any Q\&A pair or the caption is missing in the output. We include the prompt and other details on the captioning process in the supplementary material. 

Figure~\ref{fig:caption_examples} shows example images and their corresponding generated captions. While we do observe several hallucinations in the generated captions, they are also more diverse and include more details than other image-caption datasets such as SkyScript or RS5M, which are based on heuristics~\cite{skyclip} or the much simpler MLLM BLIP2~\cite{georsclip,blip} (see supplementary material for examples). Figure~\ref{fig:caption_examples} includes three examples with hallucinations to showcase their different forms. For example, the model sometimes imagines landmasses in ocean patches, man-made or water features, and provides wrong counts or length estimations.

\begin{figure}[tbh]
    \centering
    \includegraphics[width=\textwidth]{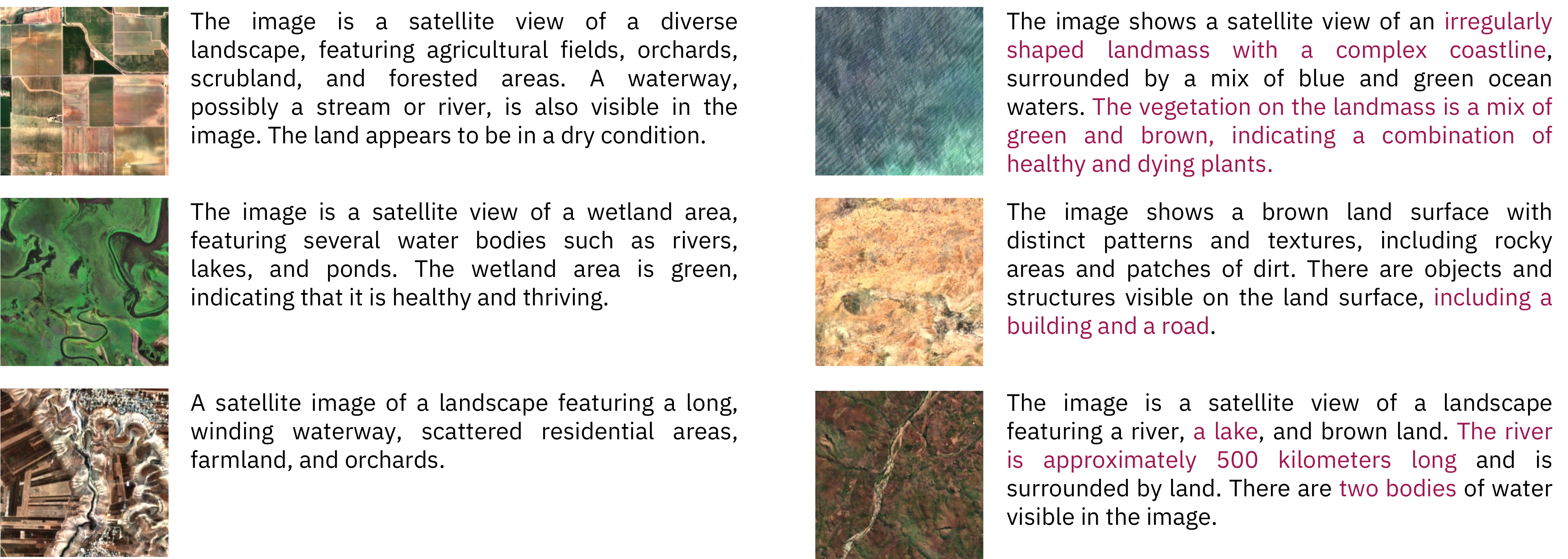}
    \caption{Examples of image-caption pairs of high quality (left) and hallucination examples (right) of our generated pre-training dataset. We highlight hallucinations in red.}
    \label{fig:caption_examples}
\end{figure}

We evaluate the captions quantitatively by comparing our validation set with manually labeled EO datasets: RSITMD~\cite{rsitmd}, RSICD~\cite{rsicd}, and UCM Captions~\cite{ucmc}. The generated captions exhibit a much higher average n-gram diversity of 0.75 compared to only 0.48 to 0.49 in the three human-annotated datasets. The similarity between captions is also lower, showing a higher lexical variety in the SSL4EO-S12-captions.

\begin{figure}[tbh]
    \centering
    \includegraphics[width=\textwidth]{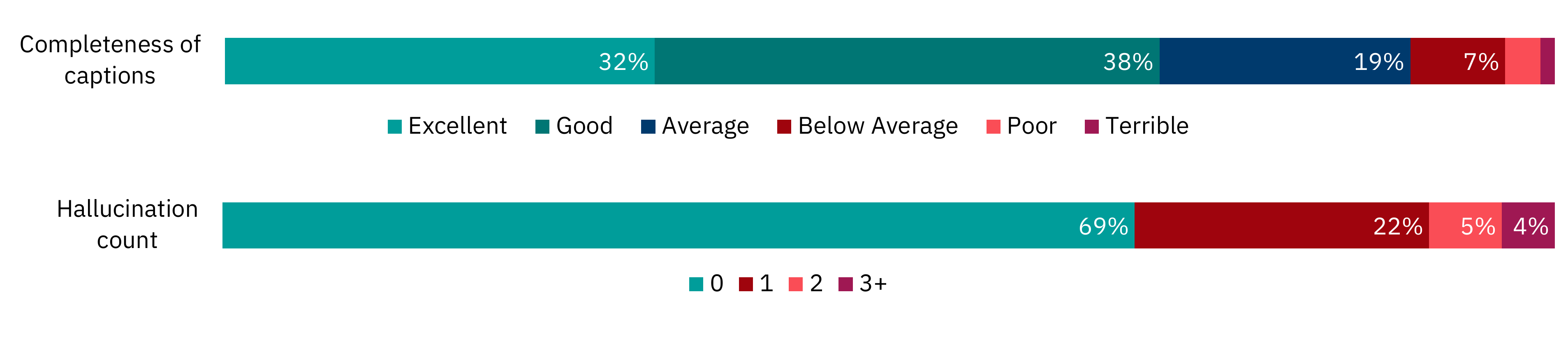}
    \caption{Evaluation results of the caption quality in the Llama3-SSL4EOS12 dataset based on 1.3k captions reviewed by domain experts. Completeness evaluates if all relevant features of an image are mentioned in the caption, while hallucinations are the number of incorrect features. }
    \label{fig:caption_quality}
\end{figure}

To assess the quality of our dataset, we asked domain experts\footnote{14 researchers from the FAST-EO project working in the Earth observation domain.} to conduct a manual evaluation using a random subset with more than one thousand captions from the validation split. Domain experts rated the captions based on completeness and presence of hallucinations. The caption completeness represents whether all relevant features in the image are covered in the caption and is measured on a scale from 0 (Terrible) to 5 (Excellent). Additionally, the experts counted hallucinations. Figure~\ref{fig:caption_quality} presents the distribution of expert ratings, indicating that over 85\% of the captions are considered to include most of the relevant features in the image. The human assessment further demonstrates that two-thirds of the evaluated data is free from hallucinations. If hallucinations are present, we typically observe only one hallucinated feature within an image. We provide details of the quantitative comparison and the human evaluation in the supplementary material.

The manual evaluation of the generated captions demonstrates that automated captioning with a general-purpose MLLM and additionally provided tags is feasible and leads to mostly correct captions. Furthermore, the quantitative assessment of the full validation set indicates that the captioning model uses a more diverse vocabulary than existing human-annotated datasets. Different from datasets like ChatEarthNet~\cite{chatearthnet} or SkyScript~\cite{skyclip} that do not use multimodal LMMs, our pipeline can capture scene-specific features like snow, clouds, or colors. While we do want to highlight the challenge of hallucinations in the dataset, our experiments show that VLMs can learn semantic concepts from the correct annotations. Furthermore, the alignment with S-1~GRD data in SSL4EO-S12 and the question-answer pairs provides additional potential for the EO community. 

\section{Llama3-MS-CLIP}

\begin{figure}[tbh]
    \centering
    \includegraphics[width=0.7\textwidth]{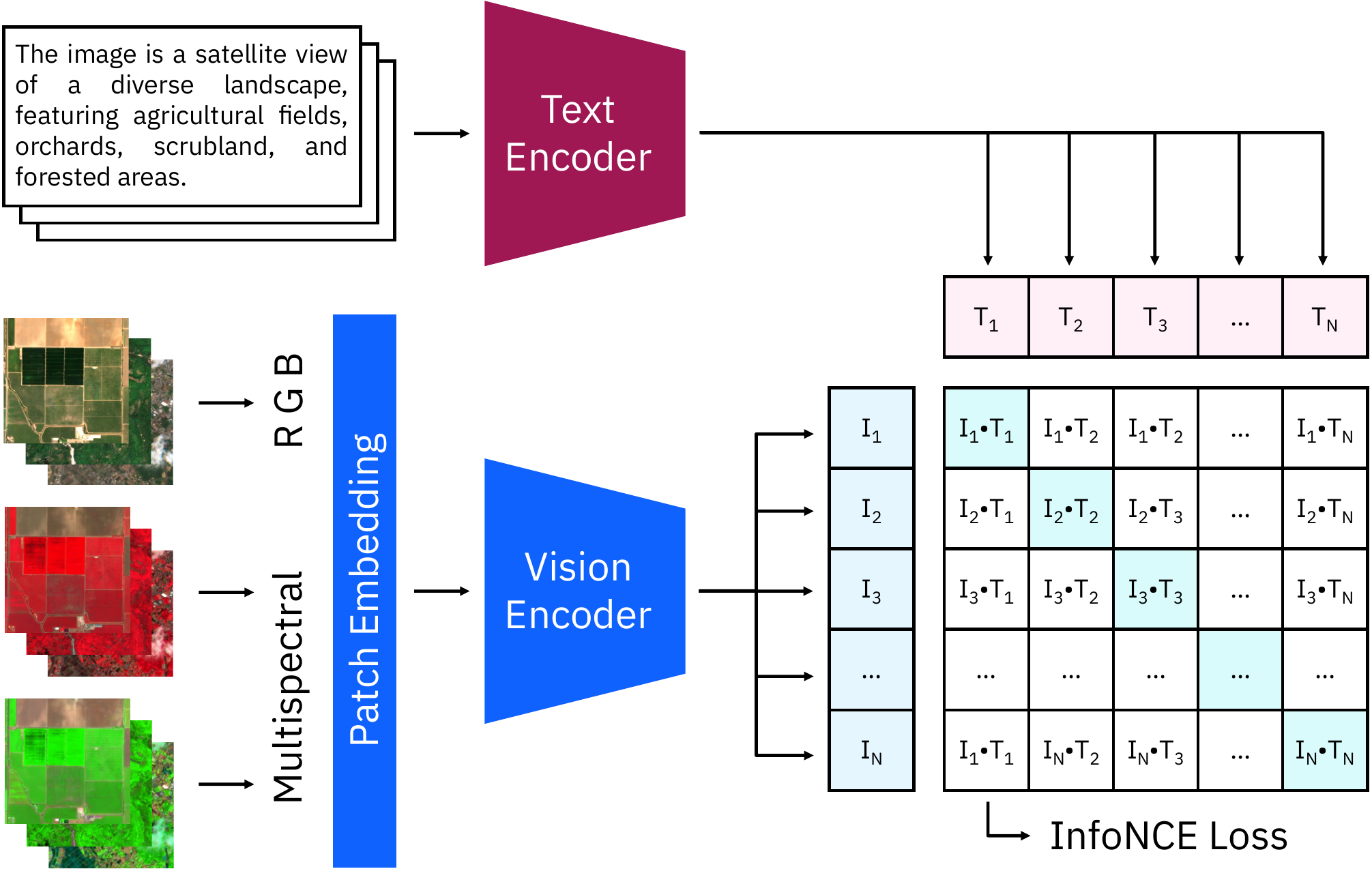}
    \caption{The CLIP model consists of two encoders for text and images~\cite{openaiclip}. We extended the RGB patch embeddings to multispectral input and initialized the weights of the additional input channels with zeros. During the continual pre-training, the images and texts of each batch are encoded and combined. The loss increases the similarity of matching pairs while decreasing other combinations.}
    \label{fig:clip}
\end{figure}

Llama3-MS-CLIP is trained with self-supervised contrastive language-image pre-training (CLIP)~\cite{openaiclip}, visualized in Figure~\ref{fig:clip}. We modified the input layer to handle Sentinel-2's spectral bands beyond RGB by extending the patch embedding for the additional channels. We initialize the corresponding weights with zero tensors so that during continual pre-training, the model starts from RGB input and can iteratively include additional channels based on optimizing the loss landscape. Hence, the model can slowly learn to leverage the additional information. Our initial experiments suggest that this initialization strategy outperforms random initialization, where the continual pre-training would be disrupted due to the noise that originates from the random weights for the additional channels.

Following CLIP, we utilize the InfoNCE loss, a contrastive loss function, to align embeddings of semantically similar samples while separating semantically dissimilar ones through cross-modal supervision. 
The InfoNCE loss encourages the embeddings of matching (positive) pairs \((x_i, y_i)\) to be similar, while pushing apart non-matching (negative) pairs \((x_i, y_j)\) with \(j \neq i\). Here, \(\mathrm{sim}(\cdot,\cdot)\) is a pairwise similarity measure, and \(\tau\) is a temperature parameter that scales the logits. Minimizing the loss thus maximizes the similarity of each positive pair relative to all negative pairs.
We summarize the training objective in equation~\ref{equ:contrastive_loss}.

\begin{equation}
\mathcal{L}_{\text{InfoNCE}} = -\frac{1}{2N}\sum_{i=1}^N \!\Bigg[\log \frac{e^{\text{sim}(x_i,y_i)/\tau}}{\sum_{j=1}^N e^{\text{sim}(x_i,y_j)/\tau}} \;+\; \log \frac{e^{\text{sim}(y_i,x_i)/\tau}}{\sum_{j=1}^N e^{\text{sim}(y_i,x_j)/\tau}}\Bigg]
\label{equ:contrastive_loss}
\end{equation}

\section{Experimental Setup}

In the following, we outline our pre-training and evaluation setting, including the downstream datasets and benchmark models.
\\

\textbf{Pre-training.}
We use the implementation and model weights provided by OpenCLIP~\cite{openclip} and perform continual pre-training of the ViT-B/16 using the SSL4EO-S12-captions. The model has 150 million parameters, split between the image and text encoders, and was initially pre-trained on LAION-2B, an English subset of the LAION-5B~\cite{schuhmann2022laionb} dataset. 
While many EO models use the ViT-B/32 version with a patch size of 32~\cite{georsclip,skyclip,remoteclip}, we find a patch size of 16 more appropriate for low-resolution images with many details. 
We used an AdamW optimizer with a learning rate of 4e-5, 50 warm-up steps, and a cosine decay scheduler that updated after each training step. The model was trained for up to five epochs on NVIDIA-A100 GPUs with a global batch size of 1200. The final model was selected based on the lowest validation loss reached after one epoch for the multispectral version and two epochs for the RGB version. Based on prior experiments, all layers are unfrozen during the pre-training of Llama3-MS-CLIP, but only the projection layers are trained for the RGB data.
\\

\textbf{Benchmark Models.} We evaluate OpenCLIP~\cite{cherti2023reproducible} ViT-B/16 and ViT-L/14 as well as three RGB-based EO-specific models based on ViT-B/32 backbones. 
SkyCLIP~\cite{skyclip} used remote sensing images with rich semantics covered in Open Street Map to construct a dataset comprising 2.6 million images and generated captions with a simple heuristic by just listing all Open Street Map tags. They performed fully unfrozen continual pre-training using the ViT B/32 backbone initialized from the LAION 2B weights by OpenCLIP~\cite{openclip}. 
RemoteCLIP~\cite{remoteclip} proposed a data scaling approach to existing datasets via annotation unification. For images with bounding box annotations, a box-to-caption generation approach was applied. The mask-to-box conversion method was used for segmentation masks. The resulting high-resolution dataset consisted of 165k images, each accompanied by five captions. RemoteCLIP is based on the OpenAI CLIP weights~\cite{openaiclip} and was adapted with fully unfrozen weights.
The authors of GeoRSCLIP~\cite{georsclip} used the images from BigEarthNet~\cite{bigearthnet} and ten other datasets. They generated captions based on the annotations and metadata using BLIP-2~\cite{blip2}. Subsequently, they fine-tuned the OpenAI ViT B/32 and ViT H/14 models applying parameter-efficient fine-tuning techniques.
\\

\textbf{Downstream Datasets.} Our zero-shot evaluation focuses on low-resolution Sentinel-2 imagery from EuroSAT~\cite{eurosat}, BigEarthNet~\cite{bigearthnet}, and METER-ML~\cite{meterml}. EuroSAT includes 64~$\times$~64 patches from ten land-use/land-cover (LULC) classes. BigEarthNet consists of S-2~L2A patches with 19 multi-labels covering a more diverse set of LULC classes and has an input size of 120~$\times$~120 pixels. Finally, METER-ML covers images with seven classes of different methane sources like \textit{landfills}, \textit{coal mines}, or \textit{natural gas processing plants}. METER-ML includes S-2 images of size 72~$\times$~72 and high-resolution RGB images from NAIP with size 720~$\times$~720. Methane is visible in the SWIR S-2 bands (bands 11 and 12) and, therefore, is an especially interesting downstream task. Since EuroSAT and METER-ML only include S-2 L1C data, we downloaded L2A data for their test sets to better align the inputs with the pre-training data. We observed improvements for METER-ML and therefore evaluated on L2A data for this task.
We perform additional experiments with the METER-ML-NAIP data and the RESISC45~\cite{resisc45} dataset. The latter includes RGB images of size 256~$\times$~256 with a spatial resolution ranging from 30\,m to 0.2\,m. The 45 scene classes range from landscapes like \textit{wetland} to large objects like \textit{airplane}. 
\\

\textbf{Evaluation.} We assess our model's zero-shot capabilities on previously unseen EO datasets. Specifically, we evaluate two tasks: zero-shot classification and text-to-image retrieval.
We adopt a template-based approach that leverages multiple prompts of the form \textit{a satellite photo of \{class name\}} and averages these for the class embedding. 
For zero-shot classification, we compute the similarity between each image and all possible class labels, assigning the class with the highest similarity score. The zero-shot classification performance is measured using macro top-1 accuracy.
We use the test set defined by CLIP for EuroSAT~\cite{openaiclip} and the official test split for all other datasets.

For the multi-label dataset BigEarthNet, we transform each class into a binary classification task. For each class, we calculate the similarity between the image embedding and the respective text embedding of that class and compare it to the mean similarity between the image and all other classes, as formulated in Equation~\ref{equ:multi-label}.
Here, $\hat{y}_i$ is the predicted label for class $i$, $x$ is the image embedding, $c_i$ is the class embedding, $\mathrm{sim}(\cdot,\cdot)$ is the dot-product similarity, and $K$ is the total number of classes.
We also compare this method to a negative-class approach (e.g., \textit{"other features"}), which boosts accuracy but substantially lowers recall and the F1~score (results in the supplementary material).

\begin{equation}
\hat{y}_i \;=\;
\begin{cases}
1, & \text{if }\mathrm{sim}\bigl(x,\,c_i\bigr) \;>\; \frac{1}{K-1}\sum_{j\neq i}\mathrm{sim}\bigl(x,\,c_j\bigr),\\[6pt]
0, & \text{otherwise}
\end{cases}
\label{equ:multi-label}
\end{equation}

For text-to-image retrieval, we calculate the similarity between a given class label and all test images, rank these scores in descending order, and then compute the mean average precision over the top 100 results (mAP@100). We report the average mAP@100 across all classes. As this retrieval procedure is class-based, it naturally extends to both single-label and multi-label datasets.

\section{Results}

We first analyze Llama3-MS-CLIP's performance and compare it against RGB-based EO models. Next, we evaluate our RGB-only model on two high-resolution tasks. Finally, we present ablation studies to investigate the effects of multispectral continual pre-training.

\begin{table}[bth] 
\centering  
\caption{Evaluation results on EuroSAT, BigEarthNet, METER-ML, and the overall average. We report zero-shot classification results in accuracy~(\%)~$\uparrow$ and text-to-image retrieval results in mAP@100~(\%)~$\uparrow$. The best-performing model is highlighted in bold, and the second-best model is underlined.}  
\label{tab:results}  
\begin{tabularx}{\textwidth}{lCCCCCCCC}  
\toprule  
& \multicolumn{4}{c}{Zero-shot classification} & \multicolumn{4}{c}{Text-to-image retrieval} \\  
\cmidrule(lr){2-5}\cmidrule(lr){6-9}  
Model & ESAT & BEN & M-ML & Avg & ESAT & BEN & M-ML & Avg \\  
\midrule
OpenCLIP B/16~\cite{openaiclip} & 39.36 & 54.28 & 38.54 & 44.06 & 48.77 & 26.70 & 24.62 & 33.36\\
OpenCLIP L/14~\cite{openaiclip} & 46.90 & 54.85 & 42.28 & 48.01 & \underline{56.92} & 28.57 & \underline{28.99} & \underline{38.16}\\
\midrule
GeoRSCLIP~\cite{georsclip}      & 52.92 & \underline{58.80} & 43.59 & \underline{51.77} & 51.36 & \underline{32.80} & 22.33 & 35.50\\
SkyCLIP~\cite{skyclip}          & 47.54 & 52.88 & \underline{44.70} & 48.37 & 53.96 & 26.29 & 28.41 & 36.22\\
RemoteCLIP~\cite{remoteclip}    & 34.02 & 52.28 & 36.42 & 40.91 & 34.34 & 26.62 & 20.08 & 27.01\\
\midrule
Llama3-RGB-CLIP                 & \underline{52.96} & 55.23 & 38.74 & 48.98 & 52.72 & 28.84 & 25.60 & 35.72\\ 
Llama3-MS-CLIP & \textbf{67.86} & \textbf{59.63} & \textbf{48.13} & \textbf{58.54} & \textbf{63.03} & \textbf{35.62} & \textbf{29.72} & \textbf{42.79} \\  
\bottomrule  
\end{tabularx}  
\end{table}

Table~\ref{tab:results} presents the zero-shot classification and retrieval results for Llama3-MS-CLIP, OpenCLIP baselines, and other RGB-based EO VLMs. Llama3-RGB-CLIP is an ablation trained using only the RGB channels of SSL4EO-S12\,v1.1.
Llama3-MS-CLIP achieves an average top-1 accuracy of 58.54\% for classification, surpassing the untuned baseline by +14.48 percentage points (pp), followed by GeoRSCLIP~\cite{georsclip} with 51.77\%. In text-to-image retrieval, Llama3-MS-CLIP outperforms all other models as well, exhibiting a 9.43pp improvement over its base model, OpenCLIP ViT-B/16. Domain-specific approaches generally outperform general-purpose baselines in zero-shot tasks, except for RemoteCLIP.

These findings underscore the effectiveness of our curated dataset and the importance of multispectral pre-training. While our RGB-based variant yields only a minor improvement over the baseline, incorporating multispectral channels leads to a substantial performance gain. Notably, GeoRSCLIP~\cite{georsclip}, despite being adapted with five times more training samples, still falls short of bridging the gap created by the missing multispectral information.

\begin{figure}[tb]
    \centering    \includegraphics[width=\linewidth]{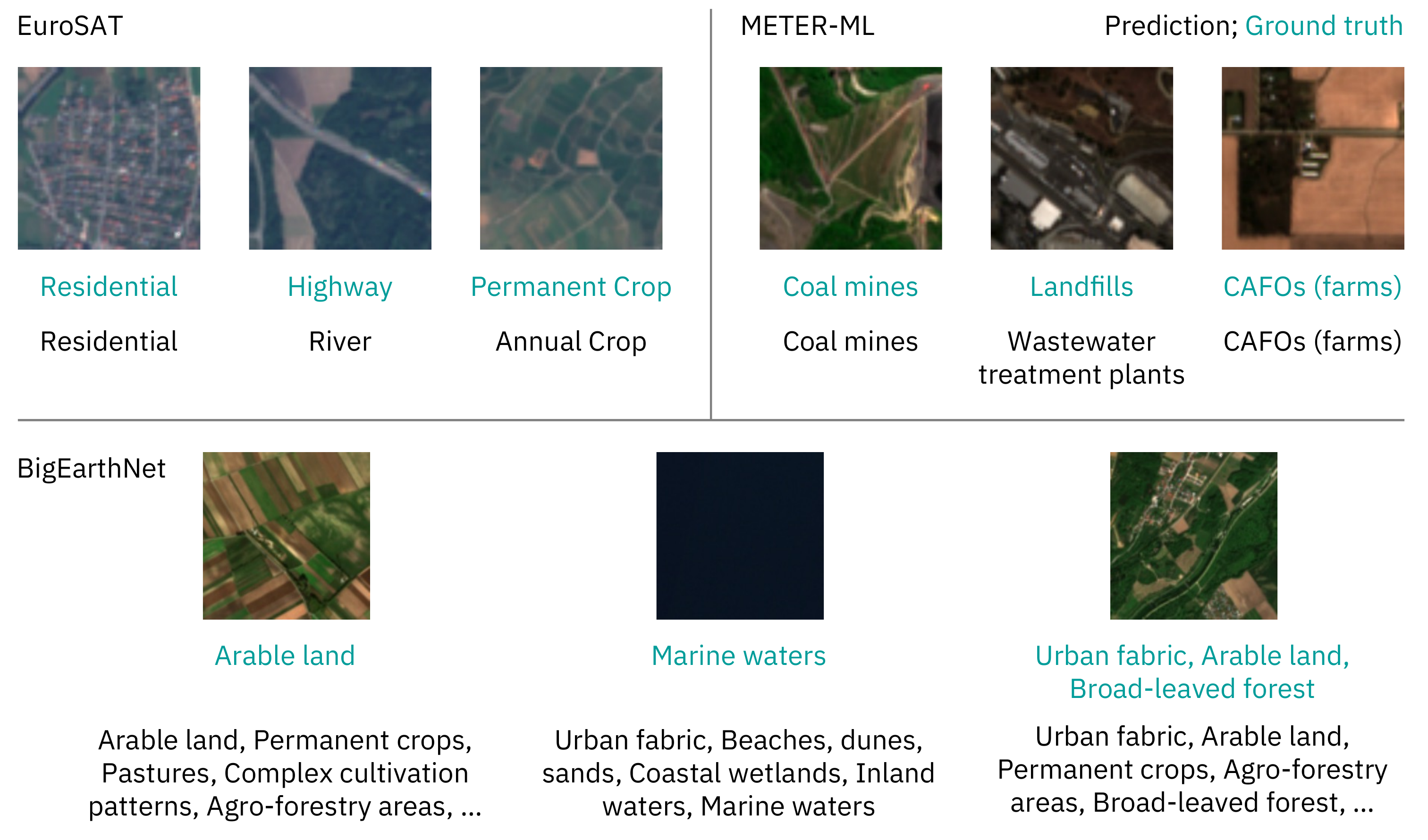}
    \caption{Prediction examples from Llama3-MS-CLIP.}
    \label{fig:predictions}
\end{figure}

We provide example predictions from Llama3-MS-CLIP in Figure~\ref{fig:predictions} to illustrate common behavior. On EuroSAT, the model is nearly always correct for general classes like \textit{residential}, \textit{sea/lake}, or \textit{industrial}, but it confuses certain pairs such as \textit{permanent crop} and \textit{annual crop}. On METER-ML, samples are often mistakenly classified as \textit{wastewater treatment plants}. In contrast, \textit{concentrated animal feedings operations} (COFAs) like farms and \textit{other features} (negative class) are mostly correctly identified.

In BigEarthNet's multi-label scenario, our approach tends to produce numerous false positives. However, these misclassifications are usually semantically correlated (e.g., \textit{inland waters} or \textit{beaches} predictions for \textit{marine waters}). Introducing an extra negative class (\textit{other features}) mitigates false positives but leads to many more false negatives. This results in higher accuracy overall, but the F1~score drops for six of the seven models. Since F1 reflects both precision and recall, we report the original method here and the alternative strategy in the supplementary material. In both cases, MS-CLIP achieves the highest F1 and significantly outperforms five other models in accuracy.

\begin{figure}[tb]
    \centering     
     \begin{subfigure}[b]{\textwidth}
         \centering
         \includegraphics[width=\textwidth]{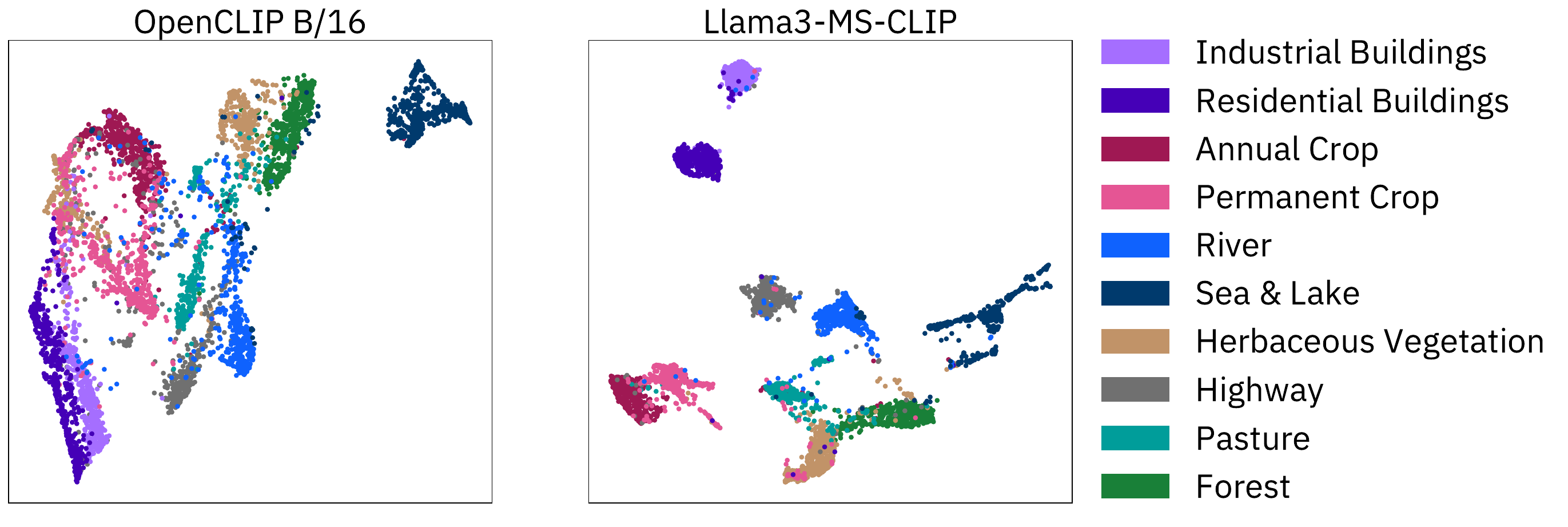}
         \caption{EuroSAT}
     \end{subfigure}
     \\
     \begin{subfigure}[b]{\textwidth}
         \centering
         \includegraphics[width=\textwidth]{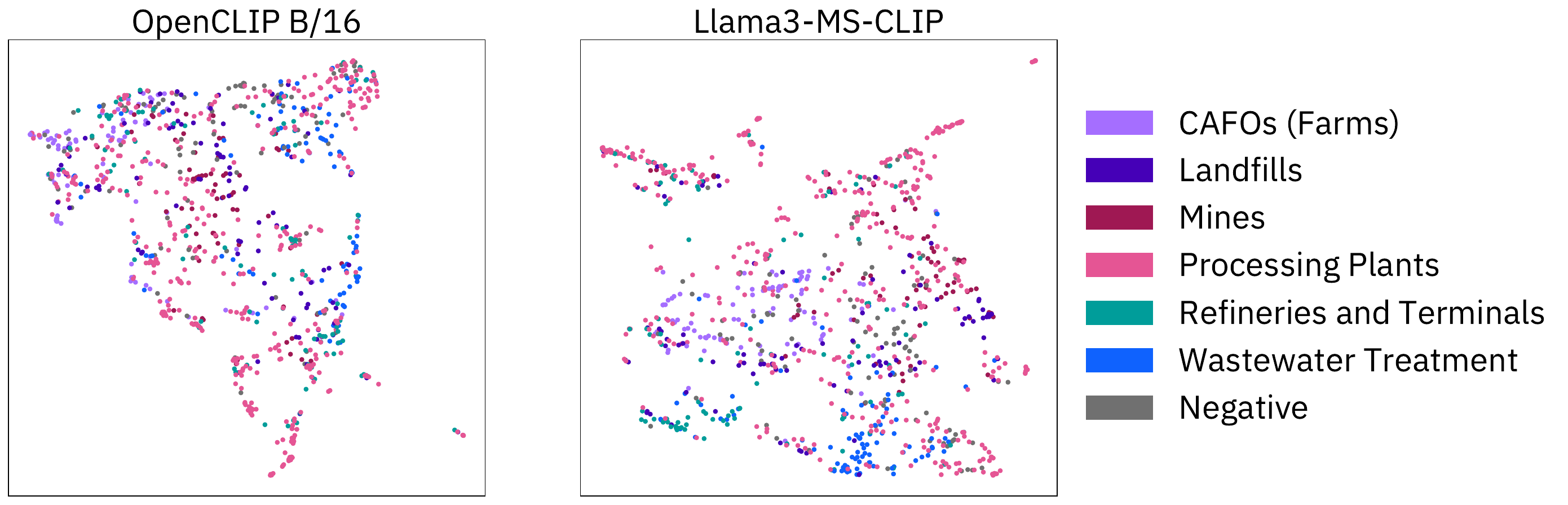}
         \caption{METER-ML}
     \end{subfigure}
    \caption{UMAP plot of the embedding spaces from OpenCLIP B/16 and Llama3-MS-CLIP for EuroSAT and METER-ML using the S-2 test images. We used a minimum distance of 0.0 for the UMAP settings with 50 and 10 nearest neighbors for EuroSAT and METER-ML, resp.}
    \label{fig:embedding_space}
\end{figure}

We further analyse the embedding space of the single-label datasets EuroSAT and METER-ML in Figure~\ref{fig:embedding_space} and compare Llama3-MS-CLIP with the base model. While many EuroSAT classes overlap in the OpenCLIP UMAP visualization, the clusters are more distinct for Llama3-MS-CLIP. Differentiating between classes in METER-ML is much more difficult, which is reflected in the lower accuracy and the embedding space visualization with no large clusters for any model. However, some classes like \textit{wastewater treatment} or \textit{refineries and terminals} form small clusters in the Llama3-MS-CLIP plot, while being more distributed for OpenCLIP.

\subsection{RGB Experiments}

We conduct additional experiments on two RGB datasets that feature high-resolution imagery, aiming to assess the generalization capabilities of our models trained solely on low-resolution Sentinel-2 data. Table~\ref{tab:high_res} shows that Llama3-RGB-CLIP achieves results on par with the untuned base model. While it scores 1.71pp lower for METER-ML-NAIP classification, it outperforms the baseline in all other tasks. GeoRSCLIP achieves the best zero-shot classification but underperforms in retrieval tasks, and RemoteCLIP again delivers the lowest results across metrics.

\begin{table}[tb]
\centering
\setlength{\tabcolsep}{2pt}
\caption{Zero-shot evaluation results for the high-resolution RGB datasets METER-ML NAIP and RESISC45. We report zero-shot classification results in accuracy (\%) $\uparrow$ and text-to-image retrieval results in mAP@100 (\%) $\uparrow$. The two best-performing models are highlighted in bold and underlined.}
\label{tab:high_res}
\begin{tabularx}{\textwidth}{lCCCCCC}
\toprule
& \multicolumn{3}{c}{Zero-shot classification} & \multicolumn{3}{c}{Text-to-image retrieval} \\
\cmidrule(lr){2-4}\cmidrule(lr){5-7}
Model & M-ML-N & RESISC45 & Avg & M-ML-N & RESISC45 & Avg \\
\midrule
OpenCLIP B/16~\cite{openaiclip} & \underline{55.09} & 67.45 & 61.27 & 37.43 & 64.30 & 50.87 \\
OpenCLIP L/14~\cite{openaiclip} & 53.48 & \textbf{72.82} & \underline{63.15} & \textbf{42.55} & \textbf{70.88} & \textbf{56.72} \\
\midrule
GeoRSCLIP~\cite{georsclip} & \textbf{59.03} & \underline{68.28} & \textbf{63.66} & 34.93 & 60.04 & 47.49 \\
SkyCLIP~\cite{skyclip} & 53.88 & 67.68 & 60.78 & \underline{39.02} & \underline{66.79} & \underline{52.91} \\
RemoteCLIP~\cite{remoteclip} & 39.35 & 66.90 & 53.13 & 23.80 & 56.88 & 40.34 \\
\midrule
Llama3-RGB-CLIP & 53.38 & 68.26 & 60.82 & 38.50 & 66.49 & 52.50 \\  
\bottomrule
\end{tabularx}
\end{table}

Comparing performance on METER-ML for both Sentinel-2 and NAIP data reveals that all models benefit from the higher resolution, with improvements ranging from 3pp to 17pp in both classification and retrieval. Our RGB variant is 5.25pp more accurate than its multispectral version in classification using S-2 imagery. Although high-resolution imagery clearly enhances VLM performance, its public availability is limited, whereas Sentinel-2 data is openly accessible every five days. Notably, relying on low-resolution data for domain adaptation does not reduce performance on high-resolution tasks, and other domain-specific approaches, even those trained on high-resolution data, only show marginal improvements over the baseline.

\subsection{Ablation Studies}

We conduct several ablation experiments to study various design choices systematically. Specifically, we investigate prompt templates, weight initialization for the patch embedding, different input bands, and strategies for freezing/unfreezing model layers.

Our initial prompt templates follow RS5M, which are already adapted for EO and outperform the original CLIP templates used for EuroSAT (see Table~\ref{tab:templates}). Extending and refining these prompts leads to a gain of on average 3.47pp in classification and 1.16pp in retrieval, relative to the baseline.

\begin{table}[tb]
\centering
\setlength{\tabcolsep}{3pt}
\caption{Zero-shot evaluation results for different text templates using the Llama3-MS-CLIP model. We report zero-shot classification results in accuracy (\%) $\uparrow$ and text-to-image retrieval results in mAP@100 (\%) $\uparrow$. The best-performing method is highlighted in bold.}
\label{tab:templates}
\begin{tabularx}{\textwidth}{lCCCCCCCC}
\toprule
& \multicolumn{4}{c}{Zero-shot classification} & \multicolumn{4}{c}{Text-to-image retrieval} \\
\cmidrule(lr){2-5}\cmidrule(lr){6-9}
Templates & ESAT & BEN & M-ML & Avg & ESAT & BEN & M-ML & Avg \\
\midrule
CLIP templates~\cite{openaiclip} & 67.64 & 60.25 & 37.33 & 55.07  & 61.14 & 34.31 & 29.45 & 41.63  \\
RS5M templates~\cite{georsclip} & 67.28 & \textbf{60.17} & 47.02 & 58.15 & 60.19 & 34.69 & \textbf{30.99} & 41.95  \\
MS-CLIP templates & \textbf{67.86} & 59.63 & \textbf{48.13} & \textbf{58.54} & \textbf{63.03} & \textbf{35.62} & 29.72 & \textbf{42.79}  \\
\bottomrule
\end{tabularx}
\end{table}

\begin{figure}[tb]
    \centering
     \begin{subfigure}[b]{0.49\textwidth}
         \centering
         \includegraphics[width=\textwidth]{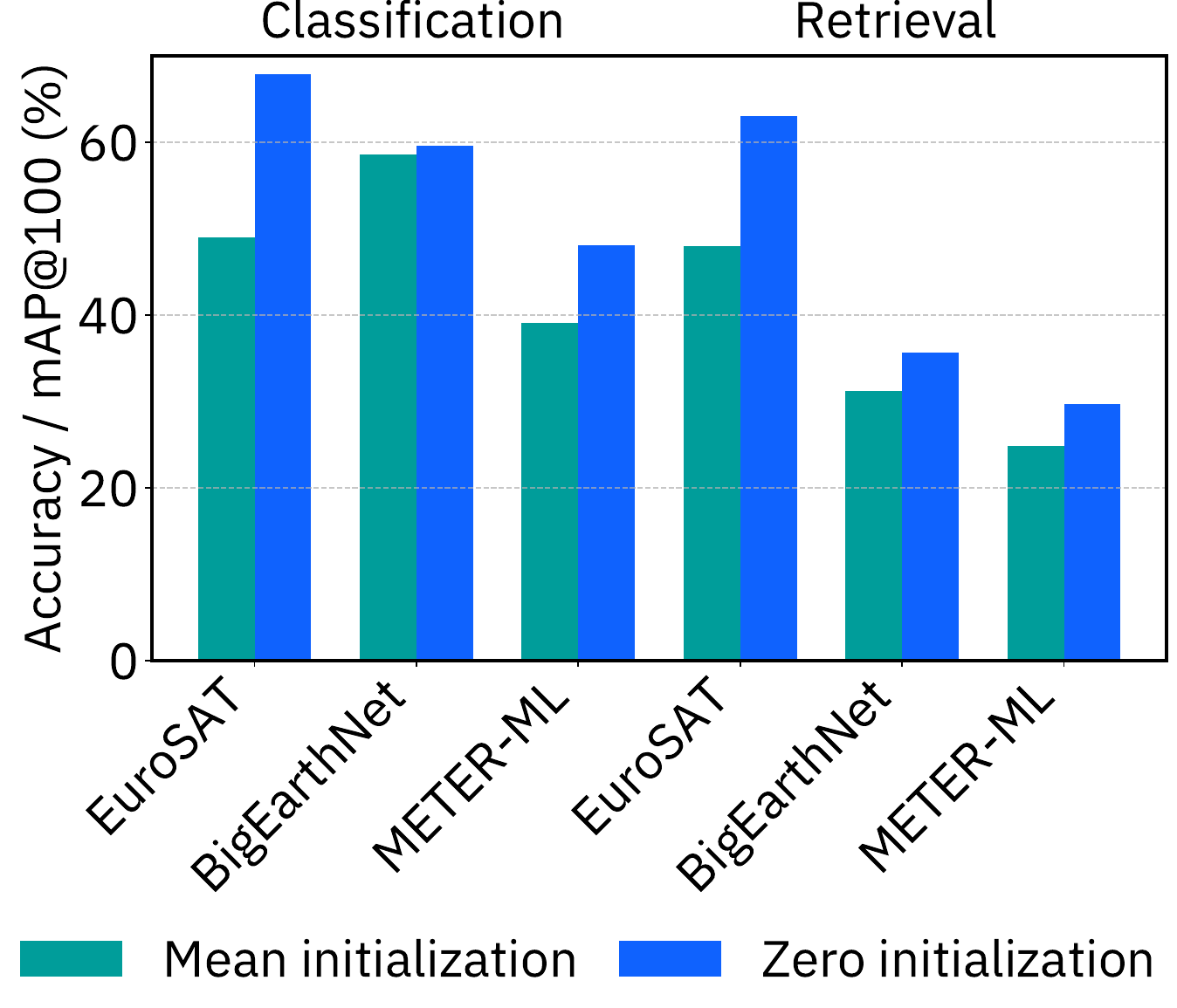}
         \caption{Patch embedding weights}
     \end{subfigure}
     \hfill
     \begin{subfigure}[b]{0.49\textwidth}
         \centering
         \includegraphics[width=\textwidth]{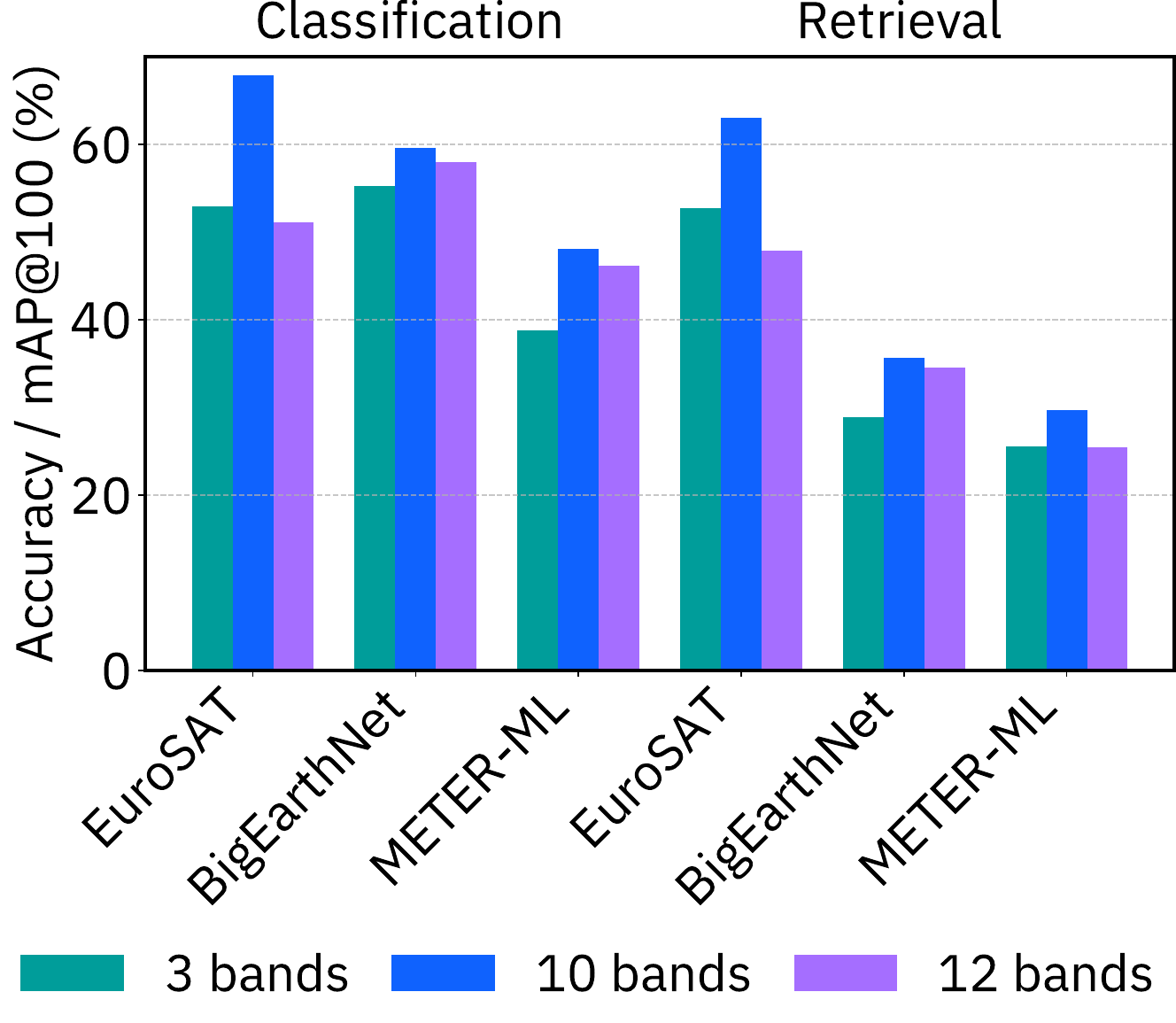}
         \caption{Input bands}
     \end{subfigure}
    \caption{Ablation experiments for the initialization of additional patch embedding channels (a) and the selected input bands (b). The final setting is in blue.}
    \label{fig:ablations}
\end{figure}

Figure~\ref{fig:ablations}(a) compares two approaches to initialize the multispectral patch embeddings. While a naive solution is to set each new channel's weights to the mean of the RGB ones, we find that zero-initialization produces superior performance in all six tasks. We hypothesize that starting from zero with a short warm-up phase lets the model adjust gradually to the additional input channels instead of the more sudden interruption with mean initialization. 

We analyse the weights of the patch embedding before and after continual pre-training. The absolute values of the patch embedding for the multispectral channels are much lower than for the RGB channels, showing that the model did not fully adapt to the additional input due to the limited number of weight updates with only one million samples. At the same time, changes in the multispectral channels are $\sim$2800 times higher than the changes in the RGB weights, which are adjusted by only 0.03\% compared to the OpenCLIP weights. This shows the additional information Llama3-MS-CLIP leverages from the multispectral channels and highlights the need for even larger multispectral EO datasets for longer pre-training without overfitting.

In Figure~\ref{fig:ablations}(b), we examine the effect of using three, ten, or all twelve Sentinel-2 bands. Although using all bands might seem beneficial, certain tasks decline in performance compared to using only the RGB bands. Omitting bands 1 and 10 and training on ten bands leads to the overall best results. The dropped bands both have a 60\,m spatial resolution, and their information might not align well with the other bands. Dropping these bands is common among EO models (e.g., in~\cite{satmae}), suggesting low meaningful information for many use cases.

We further compare various strategies for freezing and unfreezing model layers in Table~\ref{tab:unfrozen_layers}. The patch embedding was unfrozen in every setting to adapt to the multispectral input. Freezing layers can, in principle, reduce the risk of catastrophic forgetting. Yet, our experiments show that fully unfreezing the image and text encoders leads to the best results when adapting to multispectral data. By contrast, selectively fine-tuning only the projection layer yields lower performance than the baseline.
We performed a similar ablation study for Llama3-RGB-CLIP with contrary results. Fine-tuning only the projection layer resulted in the best performance. Keeping the earlier layers frozen avoids forgetting pre-trained features with RGB input, but it cannot capture the multispectral information when including the additional multispectral channels.

\begin{table}[t]
\setlength{\tabcolsep}{2pt}
\caption{Ablation experiment with different unfrozen layers. We either unfreeze the attention layers, projection layers, or all layers in the image and text encoders. We report zero-shot classification results in accuracy (\%) $\uparrow$ and text-to-image retrieval results in mAP@100 (\%) $\uparrow$. The best-performing method is highlighted in bold.}
\label{tab:unfrozen_layers}
\begin{tabularx}{\textwidth}{llCCCCCCCC}
\toprule
& & \multicolumn{4}{c}{Zero-shot classification} & \multicolumn{4}{c}{Text-to-image retrieval} \\
\cmidrule(lr){3-6}\cmidrule(lr){7-10}
Image enc. & Text enc. & ESAT & BEN & M-ML & Avg & ESAT & BEN & M-ML & Avg \\
\midrule
Attention l.  & Attention l. & 56.12 & 58.05 & 45.30 & 53.15 & 48.76 & 33.80 & 24.96 & 35.84 \\ 
Projection l. & Projection l. & 32.00 & 56.25 & 43.49 & 43.91 & 34.70 & 34.47 & 21.44 & 30.20 \\
All layers & None & 65.82 & 58.87 & \textbf{48.63} & 57.77 & 55.46 & 34.19 & 29.08 & 39.57 \\
All layers & All layers & \textbf{67.86} & \textbf{59.63} & 48.13 &\textbf{58.54} & \textbf{63.03} & \textbf{35.62} & \textbf{29.72} & \textbf{42.79}  \\
\bottomrule
\end{tabularx}
\end{table}

\section{Discussion and Limitations}

In this work, we demonstrate the benefit of leveraging MLLMs for accelerating image captioning in order to curate datasets for subsequent vision-language model training. While MLLMs are already able to digest and caption RGB representations of remote sensing images themselves, they cannot leverage multispectral data---leaving room for specialized models for Earth observation. We address this gap by coupling the automated caption generation on RGB imagery with multi-spectral data for the Llama3-MS-CLIP.
However, we acknowledge limitations when using MLLMs to generate synthetic image captions. First, we understand that there exists a risk of propagating errors or biases (e.g., in the form of hallucinations) of the MLLM further into the self-supervised models that are trained on top of the synthetic data. It will be relevant to identify such ripple effects that result from training self-supervised models on synthetically generated data of MLLMs in order to understand how errors and biases are propagated, reduced, or reinforced by downstream models. Second, we note that the existing dataset likely benefits from increased diversity, as the word count graph in the supplementary section highlights a trend of similar topics in many of the captions. Third, we advocate for considering human-in-the-loop systems during the caption generation process in settings where errors by the MLLMs are not acceptable.

Based on the synthetically generated captions, Llama3-MS-CLIP demonstrates the benefit of using multi-spectral data during pretraining. Even though we observe significant performance improvements when leveraging multi-spectral data instead of RGB data, our experiments also show that the model is potentially not yet saturated. This is indicated by comparably low weights in the patch embedding of Llama3-MS-CLIP for the non-visible channels compared to the visible RGB spectrum. Overall, this experiment reinforces our expectation that the performance of the model will further improve with longer continuous pretraining, leveraging a larger and more diverse training corpus. 

Finally, we see a possibility for future research to work on integrating and merging pixel-level training strategies with the image-level training we employ in this work. This merge could improve the model's capability to capture detailed image nuances and help differentiate between closely related classes. Pixel-level understanding might also unlock additional progress on other tasks that we did not explore in this work, including semantic segmentation and object detection.

\section{Conclusion}
We introduce a multispectral vision-language dataset of low-resolution, multispectral Sentinel-2 data with corresponding captions. Our automated captioning strategy scales easily, reducing the need for costly human annotations. 
On top of this dataset, we build Llama3-MS-CLIP, the first CLIP-like multispectral VLM. Our experiments show that our model significantly improves zero-shot classification and retrieval compared to other methods, even domain-specific adaptations trained on larger datasets. We see significant potential in leveraging the open-sourced Llama3-MS-CLIP in downstream applications and as a vision encoder for building multispectral MLLMs.

\begin{credits}
\subsubsection{\ackname} 
We thank the remote sensing experts for reviewing the generated captions and Niklas Kopp for providing the embedding space analysis.

\subsubsection{\discintname}
This work is part of the FAST-EO project funded by the European Space Agency (ESA), contract number 4000143501/23/I-DT.

\end{credits}


\clearpage
\setcounter{page}{1}

\title{Beyond the Visible: Multispectral Vision-Language Learning for Earth Observation}

\titlerunning{Multispectral Vision-Language Learning for Earth Observation}

\author{Marimo, C., Blumenstiel, B., Nitsche, M., Jakubik, J., and Brunschwiler, T.}

\institute{\large{Supplementary Material}}

\maketitle

We provide additional information regarding the captioning, caption analysis, data processing, and some additional model evaluations.

\section{Captioning}

We first provide our approach of selecting the captioning model, followed by a more detailed quantitative analysis of the generated captions.

\begin{table}[tbh]
	\centering
    \caption[METEOR Scores for different image captioning models]{METEOR scores of different image captioning models on EO image-caption datasets.}
	\label{tab:meteor}
	\begin{tabularx}{\textwidth}{lCCCC}
		\toprule
		\textbf{Model} & \textbf{RSITMD} & \textbf{UCM Captions} & \textbf{RSICD} & \textbf{Average}  \\
		\hline
		 BLIP2~\cite{blip2}           & 0.08           & 0.12   & 0.10  & 0.10     \\
		
		Llama3-LLaVA-Next-8B~\cite{llavanext}  &\textbf{0.20}          &\textbf{ 0.25 }  &0.15 &\textbf{0.20  }       \\

		RS-LLaVA~\cite{rsllava}           & 0.14         & 0.20     & \textbf{0.16 }& 0.16     \\
        \bottomrule
        
	\end{tabularx}
	
\end{table}

\begin{figure}[tbh]
    \centering
    \includegraphics[width=\textwidth]{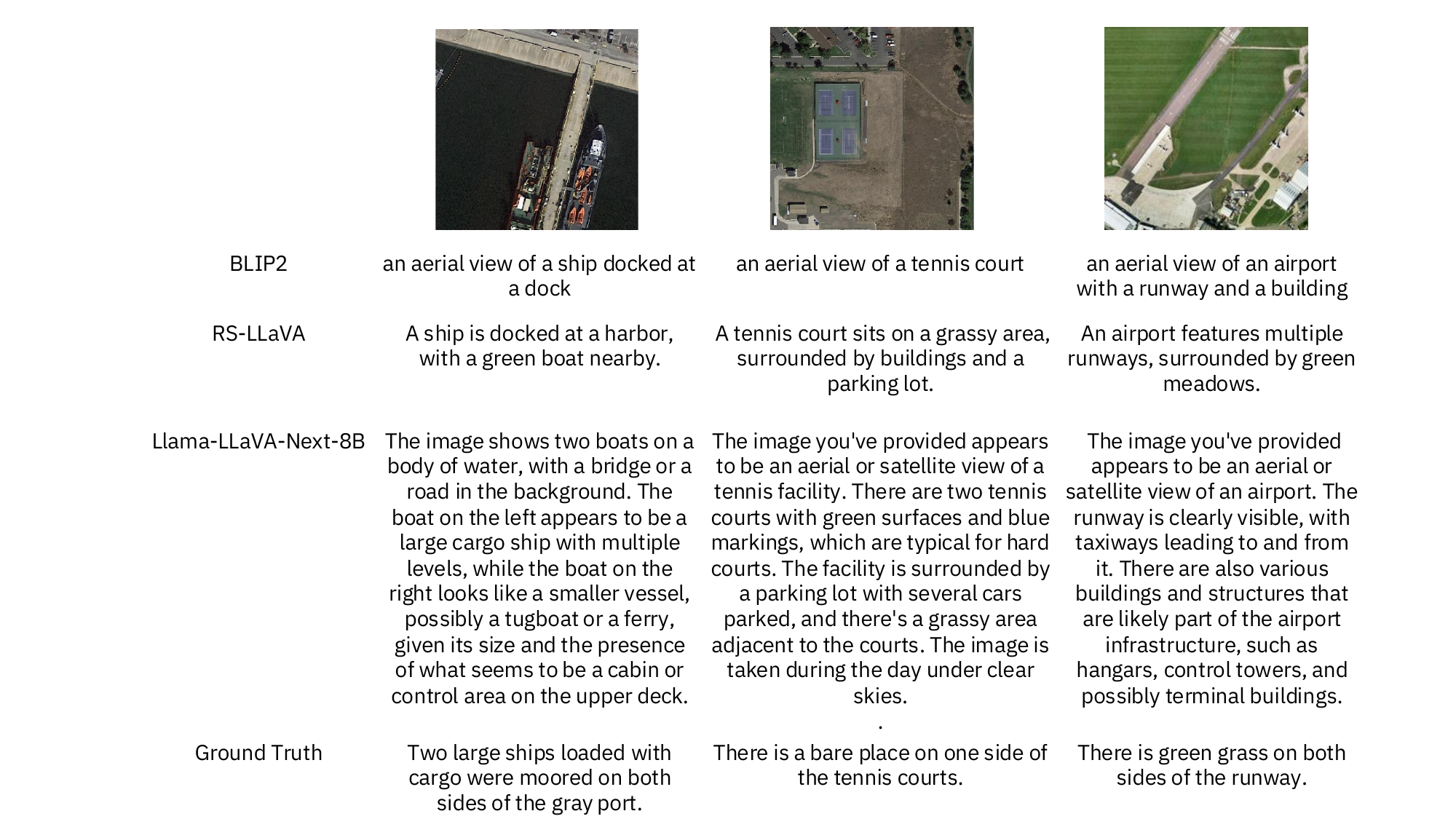}
    \caption{Example captions from the different captioning models at the same settings. Llama3-LLaVA-Next-8B provides more detailed captions, which are closer to the ground truth.}
    \label{fig:model_test}
\end{figure}

We applied the METEOR (Metric for Evaluation of Translation with Explicit Ordering)~\cite{meteor} to evaluate three captioning models: BLIP2~\cite{blip2}, utilized for generating captions for the RS5M~\cite{georsclip} dataset, Llama3-LLaVA-Next-8B~\cite{llavanext}, and RS-LLaVA~\cite{rsllava}, a LLaVA model specifically finetuned on remote sensing data. METEOR accounts for synonyms and paraphrases when aligning the generated text with reference texts. We assessed these models using the UCM Captions~\cite{ucmc}, RSICD~\cite{rsicd}, and RSITMD~\cite{rsitmd} datasets, all of which provide ground truth captions for images, enabling objective scoring by comparing machine-generated captions with the ground truth. The models were implemented through the Huggingface Transformers library. As detailed in Table~\ref{tab:meteor}, the Llama3-LLaVA-Next-8B model outperformed the others with an average score of 0.2. 
Figure~\ref{fig:model_test} presents sample captions from each of the three models across three different images, each sourced from one of the three datasets. Llama3-LLaVA-Next-8B generates more comprehensive captions, whereas BLIP2 and RS-LLaVA offer succinct descriptions with identical configurations.

Based on the METEOR scores detailed in Table \ref{tab:meteor}, we selected the Llama3-LLaVA-Next-8B model as our captioning model. We incorporated descriptive tags of the image into the final prompt and assigned the model the role of a remote sensing expert. The model was instructed to deliver output in a chain-of-thought procedure by initially formulating three questions and answers concerning the image, followed by summarizing the image in an informative and objective caption. The captioning prompt is provided below:

\begin{tcolorbox}[boxrule=0pt, sharp corners, width=0.98\textwidth, boxsep=7pt, left=5pt, right=5pt, top=5pt, bottom=5pt]
\textit{You are a remote sensing expert. Your task is to write a set of 3 questions and answers that require another expert to observe the satellite image in order to answer them. The questions must be diverse, creative, and relate to different aspects such as object attributes, object positions and relationships, or weather observations. Answers must be informative and objective. Then summarize the image in an informative caption. You must not interpret the image, but only describe what you see. You must not imagine water features, they are always listed if they are present. You must not ask for the sky. 
}\\
\textit{
Use the following keywords on all images QUESTION:, ANSWER:, and CAPTION: and make sure these keywords appear in your output. 
}\\
\textit{
The image includes the following features, among others: \{tags\}. 
}\\
\textit{
Places in the image include: \{names\}.
}\end{tcolorbox}

The prompt is provided along with the RGB image.
While running Llama3-LLaVA-Next-8B, we used a temperature of 0.3, top$_{k}$ of 0.9, and a top$_{k}$ of 50 to generate more descriptive captions rather than creative ones. The sentences about tags and names are only included when they are present for a given image.

\begin{figure}[ht]
    \centering
     \begin{subfigure}[b]{0.45\textwidth}
         \centering
         \includegraphics[width=\textwidth]{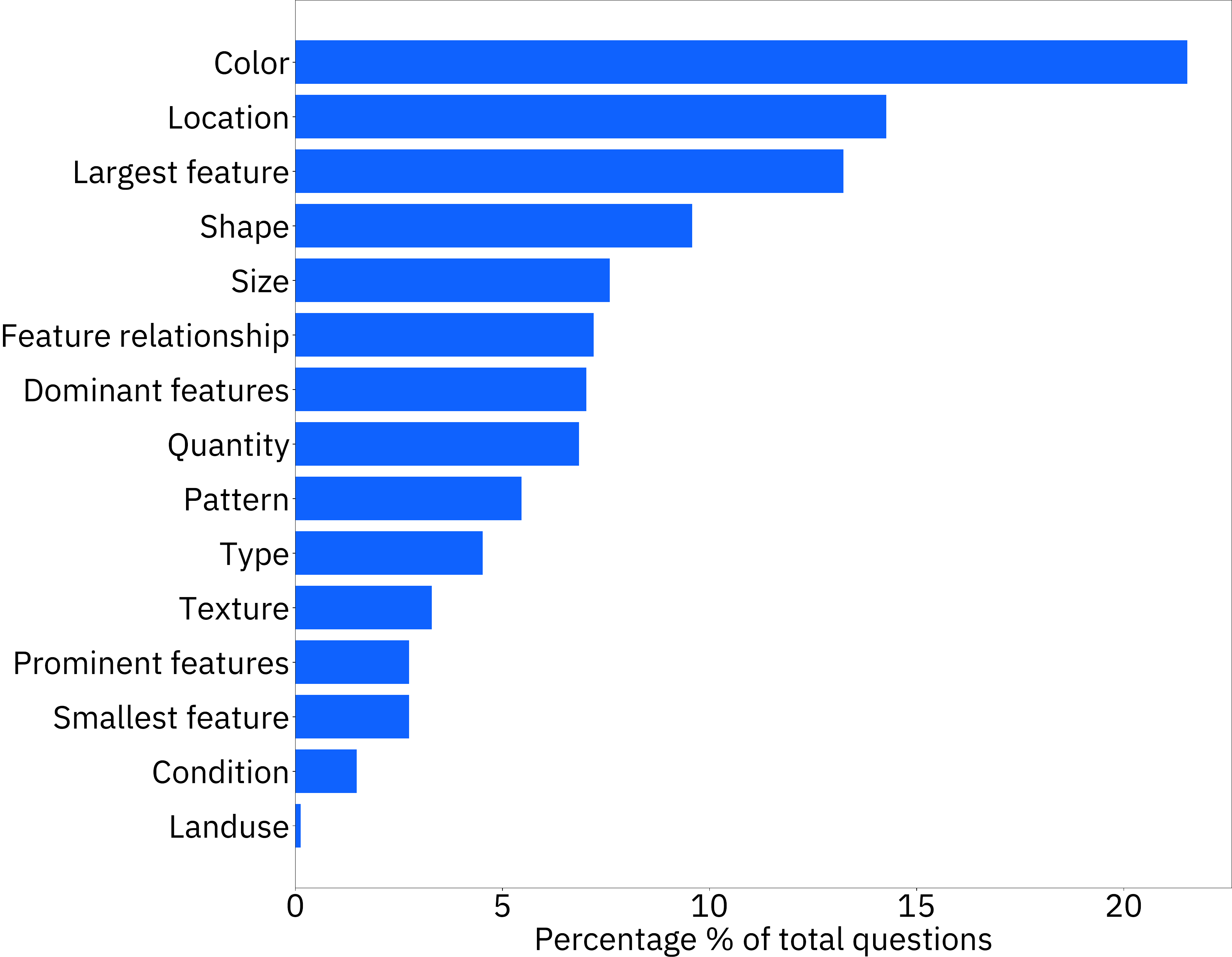}
         \caption{Question type}
     \end{subfigure}
     \hfill
     \begin{subfigure}[b]{0.45\textwidth}
         \centering
         \includegraphics[width=\textwidth]{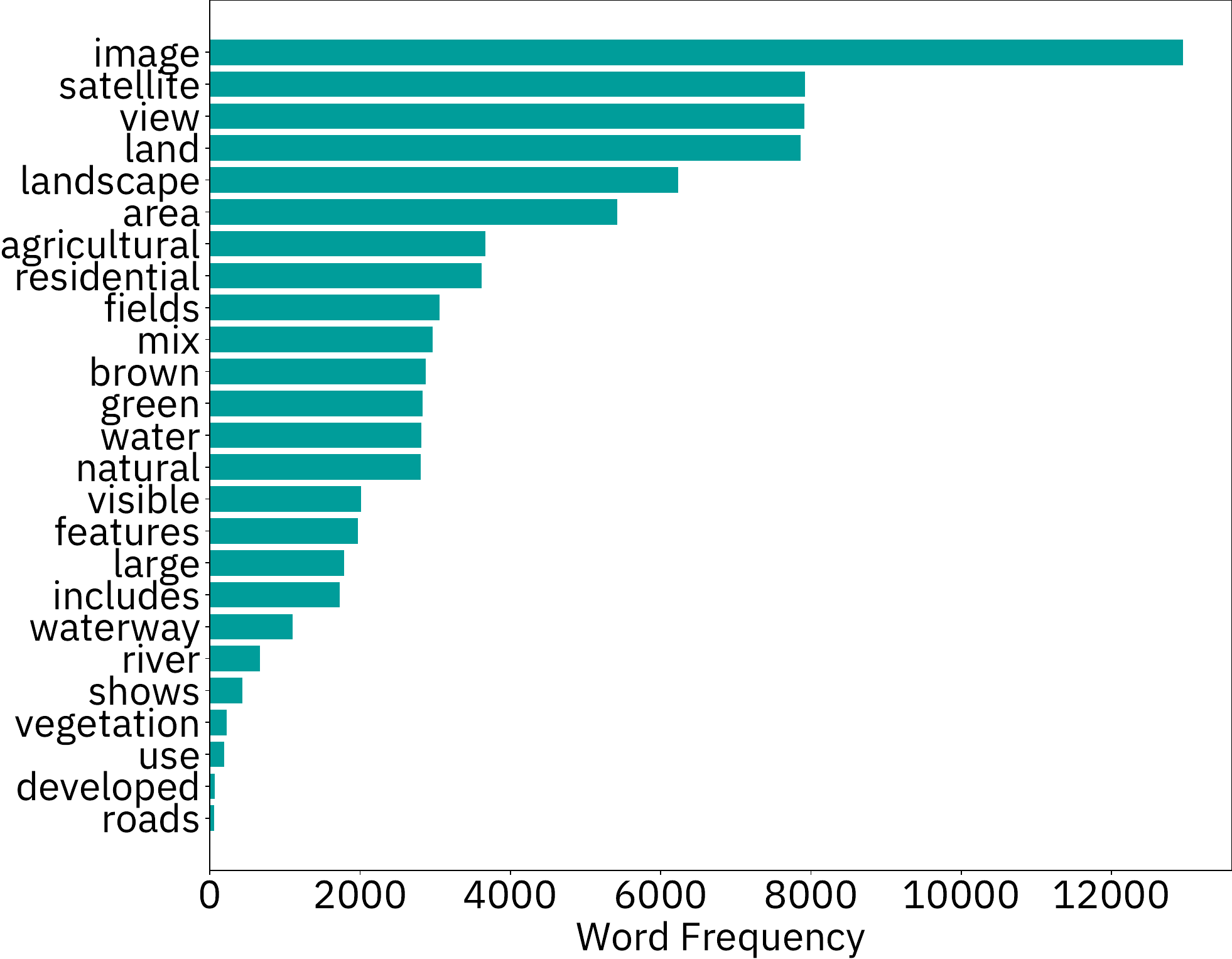}
         \caption{Word frequency}
         \label{fig:three sin x}
     \end{subfigure}
    \caption{Question type and word frequency distribution in the generated Q\&A pairs and captions of the validation split.}
    \label{fig:word_distribution}
\end{figure}

\begin{figure}[tbh]
    \centering
    \includegraphics[width=0.6\textwidth]{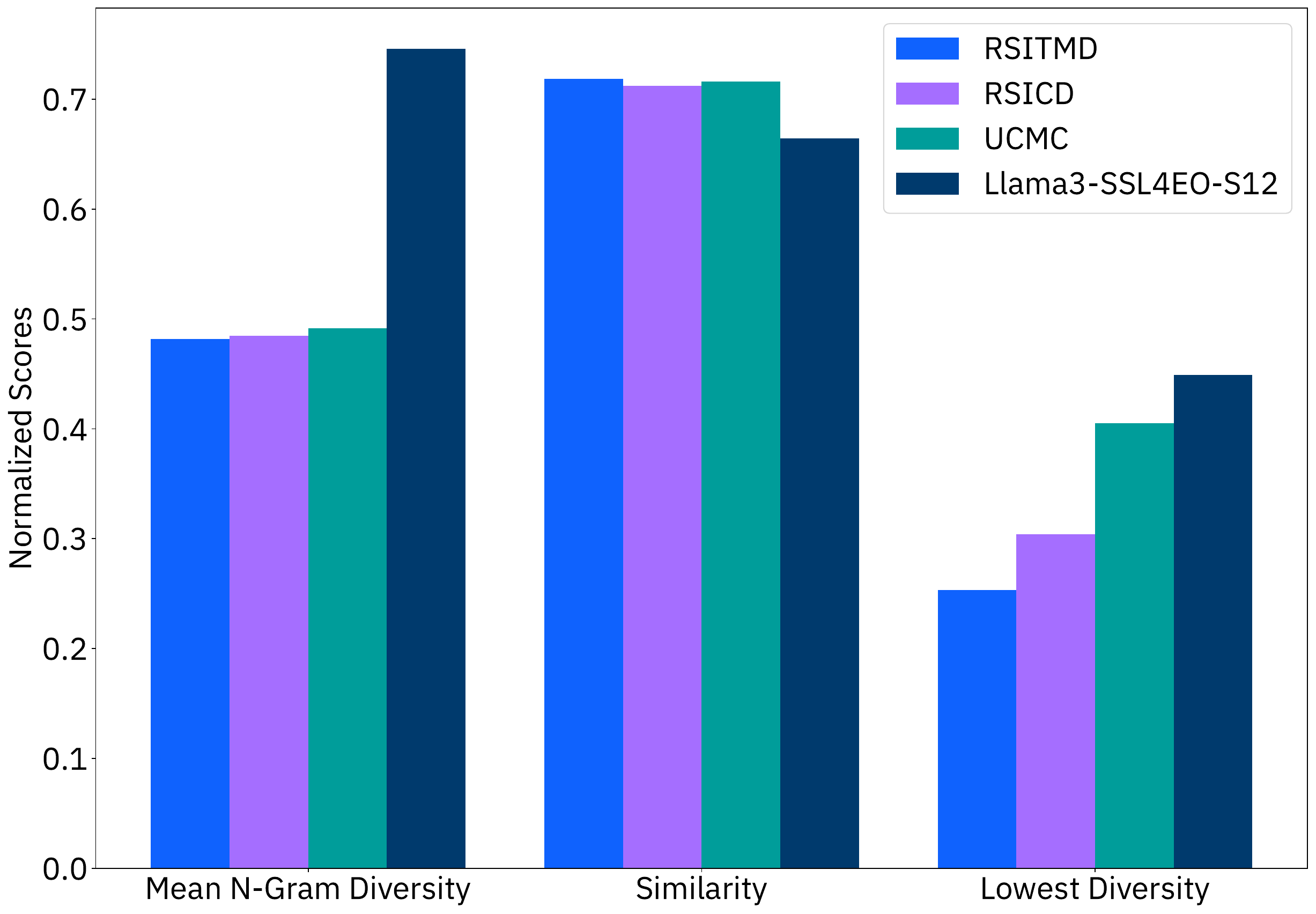}
    \caption{Caption diversity of different datasets on the validation split. The diversity is measured per caption, and similarity is measured between all captions.}
    \label{fig:diversity}
\end{figure}

\begin{figure}[tbh]
    \centering
    \includegraphics[width=\textwidth]{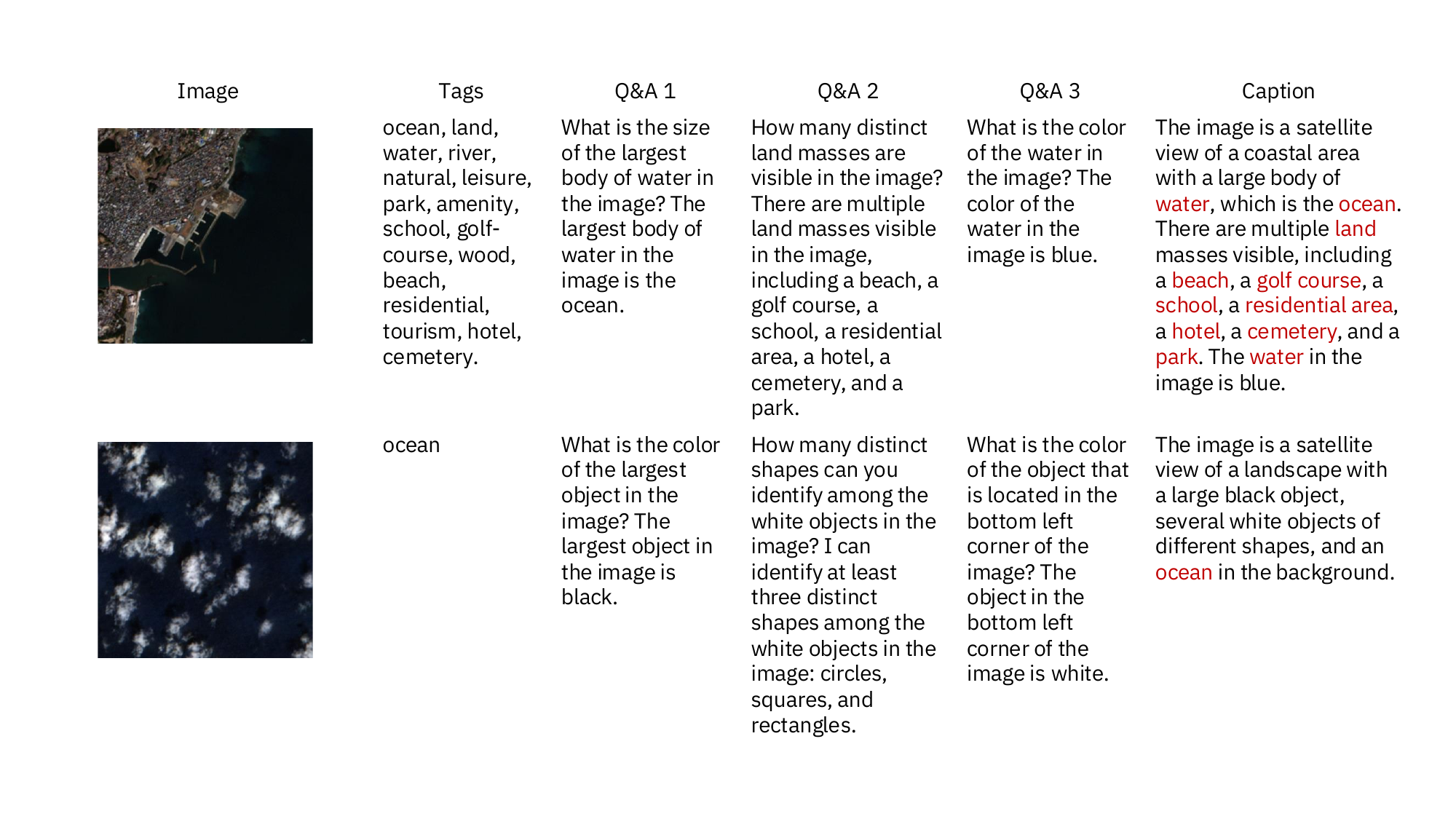}
    \caption{Example output from the captioning pipeline including the tags, questions, and captions.}
    \label{fig:example_cap}
\end{figure}

The leading question types and the most frequent words are displayed in Figure~\ref{fig:word_distribution}, highlighting the dataset's diversity. We also evaluated the lexical and n-gram diversity of the captions to assess the semantic diversity present in the data. Compared to typical EO image-caption datasets on the validation split, our dataset exhibited higher average n-gram diversity and lexical variety, as demonstrated in Figure~\ref{fig:diversity}. Example output of the captioning pipeline is shown in Figure~\ref{fig:example_cap}.

\begin{figure}[tbh]
    \centering
    \includegraphics[width=\textwidth]{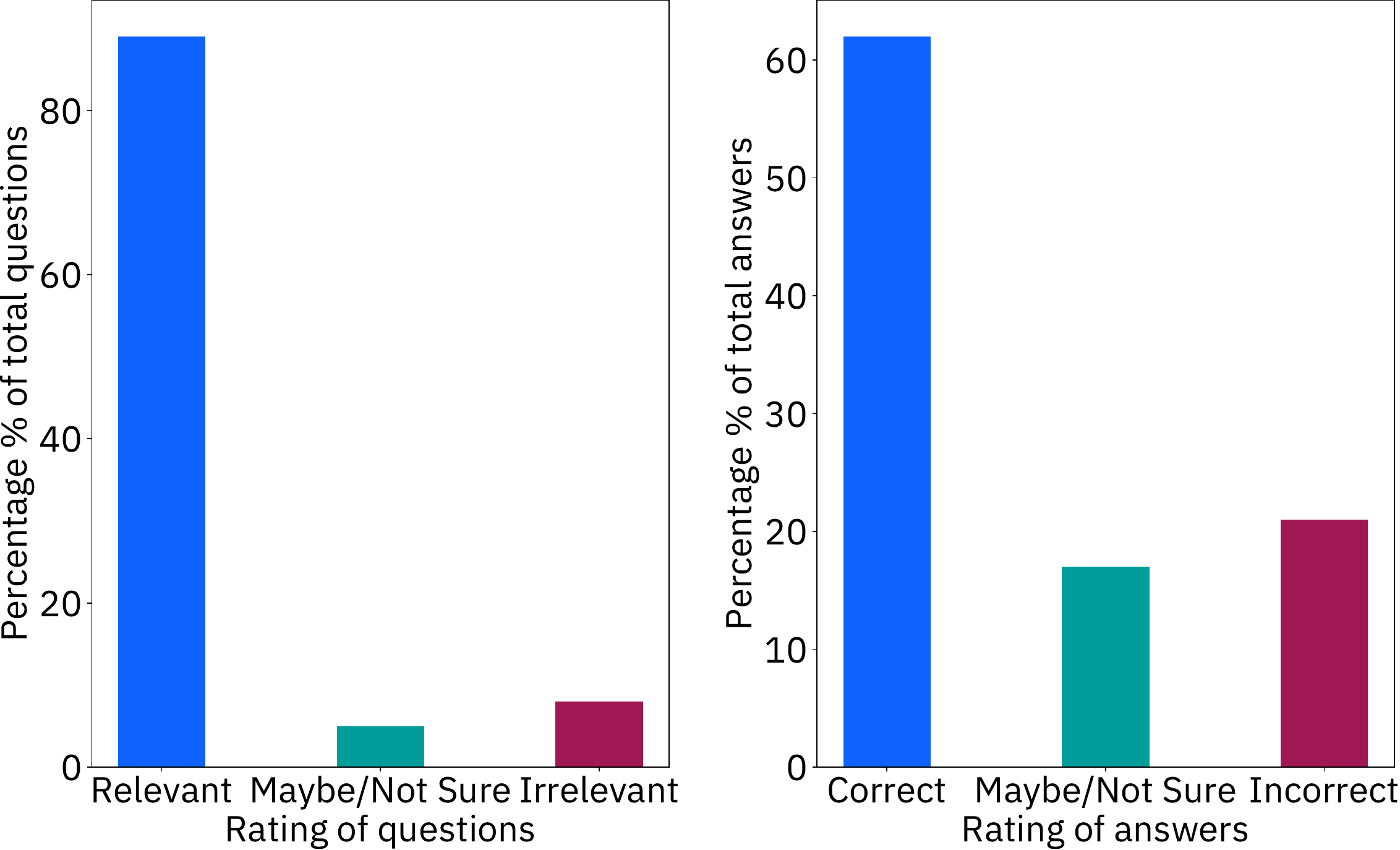}
    \caption{Question and answer rating by human experts on the Llama3-SSL4EO-S12 dataset}
    \label{fig:qa_ratings}
\end{figure}

We engaged human experts to assess a subset of over one thousand captions, employing a level rating system to judge the correctness of the captions. 
In addition to the evaluation presented in the main paper, we also evaluated the questions and answers.
The questions were judged based on relevancy, while the answers were evaluated based on their correctness relative to the question. 
Questions were rated as either \textit{relevant}, \textit{irrelevant}, or \textit{maybe/not sure}. Answers were rated as either \textit{correct}, \textit{incorrect}, or \textit{maybe/not sure}. Figure~\ref{fig:qa_ratings} illustrates the statistics. Approximately 85\% of the questions were rated as relevant for the given image, while more than 80\% of the answers were rated as either correct or maybe correct.

\section{Data processing}

We shortly outline any data processing steps for the downstream tasks.
The zero-shot performance can vary significantly based on the exact label text~\cite{mess}. We therefore keep the original class names for EuroSAT and BigEarthNet. We write out abbreviations for METER-ML and use \textit{other features} for the negative class.

For all models, the input images were resized to 224~$\times$~224 pixels using bicubic interpolation. For RGB models, we selected the RGB bands from the Sentinel-2 data and scaled the reflectance range 0––2000 to the 0––255 range of unit8. This is the typical range in literature~\cite{ssl4eos12_v11} for RGB values. Clouds or snow have higher values, but are typically not present in the downstream datasets. Finally, normalization was applied using the mean and standard deviation values used during pre-training of each model.

\section{Additional results}

\begin{table}[tb]
\centering
\caption{F1 scores (\%) $\uparrow$ for zero-shot classification. The best-performing model is highlighted in bold, and the second-best model is underlined.}
\label{tab:f1}
\begin{tabularx}{\textwidth}{lCCCC}
\toprule
\textbf{Model} & \textbf{EuroSAT} & \textbf{BigEarthNet} & \textbf{METER-ML} & \textbf{Average} \\
\midrule
OpenCLIP B/16 \cite{openclip} & 40.91 & 33.33  & 14.24 & 29.49  \\
OpenCLIP L/14 \cite{openclip} & 47.35 & \underline{34.58} & 23.27 & 35.06  \\
\midrule
GeoRSCLIP \cite{georsclip} & 47.04 & 34.40 & 11.98 & 31,14  \\
SkyCLIP \cite{skyclip} & 48.50 & 32.42 & \underline{25.72} & \underline{35.54} \\
RemoteCLIP \cite{remoteclip} & 27.34 & 30.67 & 17.49 & 25.16  \\
\midrule
Llama3-RGB-CLIP & \underline{53.58} & 33.41& 17.08 & 34.69  \\ 
Llama3-MS-CLIP & \textbf{64.27} & \textbf{36.87} & \textbf{32.27} & \textbf{44.47}  \\ 
\bottomrule
\end{tabularx}
\end{table}

We provide some additional metrics from our experiments. 
Table~\ref{tab:f1} presents the F1~scores for the three downstream datasets. Llama3-MS-CLIP has the highest F1 scores among all datasets, with an 8.93\% difference to SkyCLIP, the second-best model. This highlights that the model balances precision and recall with fewer false positives and false negatives.

\begin{table}[tbh]
\centering  
\caption{Various metrics (\%) for zero-shot classification on BEN using two different methods for multilabel classification.}  
\label{tab:ben_class_methods}  
\begin{tabularx}{\textwidth}{lCCCCCCCC}  
\toprule  
& \multicolumn{4}{c}{Average over other classes} & \multicolumn{4}{c}{"Other Features" class} \\  
\cmidrule(lr){2-5}\cmidrule(lr){6-9}  
Model & Acc. & Prec. & Recall  & F1 & Acc.  & Prec. & Recall & F1 \\  
\midrule  
OpenCLIP B/16 \cite{openaiclip} & 54.28 &23.44  &74.77 & 33.33 & \textbf{78.10} & 26.48 & 24.21 & 21.33 \\  
OpenCLIP L/14 \cite{openaiclip} & 54.85 & \underline{26.15} & \underline{75.58} & \underline{34.58} & 56.03 & 32.21 & \underline{42.58} & 27.72 \\  
\midrule  
GeoRSCLIP \cite{georsclip}  & \underline{58.80} & 24.55 & 77.23 & 34.40 & 60.94 & \textbf{38.57} & 30.59 & \underline{30.29} \\  
SkyCLIP \cite{skyclip}  & 52.88 & 25.23 & 73.09 & 32.42 & 54.13 & 26.86  & 41.93 &  24.62\\  
RemoteCLIP \cite{remoteclip} & 52.28 & 21.66 & 73.06 & 30.67 & 53.32 & 25.47 & 41.89 & 27.65 \\  
\midrule  
Llama3-RGB-CLIP & 55.23 & 23.52 & 74.82 & 33.41 & 57.19 & 29.60  & 29.32 & 24.18 \\  
Llama3-MS-CLIP & \textbf{59.63} & \textbf{26.68} & \textbf{77.84} & \textbf{36.87} & \underline{70.73} & \underline{34.09} & \textbf{57.06} &\textbf{38.74}  \\  
\bottomrule  
\end{tabularx}  
\end{table}

To perform zero-shot classification on multilabel datasets, we experiment with two methods. Both transfer the task into binary classification tasks for each class.
The first method uses the remaining classes as a negative prototype. For each class, the similarity scores between the text embeddings of all classes except the current one and the image embedding are averaged.
This average is used as a threshold to determine if the current class (which is not included in the threshold) is being predicted by the model or not. Because some classes are closely related to the target classes, this results in more false positives.
The second method introduces an extra class called \textit{other features}. A class is predicted if the similarity score between its text embedding and the image is higher than this negative label.
Table \ref{tab:ben_class_methods} shows the metrics of using the two methods.

\begin{table}[tb]
\centering
\setlength{\tabcolsep}{3pt}
\caption{Ablation for the weight initialization of the additional multispectral patch embedding layers comparing zero weights with the mean of the pre-trained RGB channels. We report zero-shot classification results in accuracy (\%) $\uparrow$ and text-to-image retrieval results in mAP@100 (\%) $\uparrow$. The best-performing method is highlighted in bold.}
\label{tab:patch_init}
\begin{tabularx}{\textwidth}{lCCCCCCCC}
\toprule
& \multicolumn{4}{c}{Zero-shot classification} & \multicolumn{4}{c}{Text-to-image retrieval} \\
\cmidrule(lr){2-5}\cmidrule(lr){6-9}
Method & ESAT & BEN & M-ML & Avg & ESAT & BEN & M-ML & Avg \\
\midrule
Mean initialization & 49.00 & 58.59 & 39.05 & 48.88  & 47.96 & 31.26 & 24.87 & 34.69\\
Zero initialization & \textbf{67.86} & \textbf{59.63} & \textbf{48.13} & \textbf{58.54} & \textbf{63.03} & \textbf{35.62} & \textbf{29.72} & \textbf{42.79} \\
\bottomrule
\end{tabularx}
\end{table}

\begin{table}[tb]
\centering
\setlength{\tabcolsep}{2pt}
\caption{Zero-shot evaluation with varying input bands, specifically the Sentinel-2 RGB bands, all ten bands with 10\,m or 20\,m resolution, and all 12 S-2 bands. We report zero-shot classification results in accuracy (\%) $\uparrow$ and text-to-image retrieval results in mAP@100 (\%) $\uparrow$. The best-performing method is highlighted in bold.}
\label{tab:input_bands}
\begin{tabularx}{\textwidth}{lCCCCCCCC}
\toprule
& \multicolumn{4}{c}{Zero-shot classification} & \multicolumn{4}{c}{Text-to-image retrieval} \\
\cmidrule(lr){2-5}\cmidrule(lr){6-9}
Input bands & ESAT & BEN & M-ML & Avg & ESAT & BEN & M-ML & Avg \\
\midrule
3 S-2 bands (RGB) & 52.96 & 55.23 & 38.74 & 48.97 & 52.72 & 28.84 & 25.60 & 35.72 \\ 
10 S-2 bands &  \textbf{67.86} & \textbf{59.63} & \textbf{48.13} & \textbf{58.54} & \textbf{63.03} & \textbf{35.62} & \textbf{29.72} & \textbf{42.79}\\ 
12 S-2 bands & 51.14 & 58.00 & 46.21 & 51.78 & 47.83 & 34.55 & 25.50 & 35.96 \\ 
\bottomrule
\end{tabularx}
\end{table}

Table~\ref{tab:patch_init} and~\ref{tab:input_bands} provide the results for the ablation studies presented in the main paper related to the initialization and the input bands.
To incorporate multispectral input, it is necessary to modify the image encoder's patch embedding by expanding the channels and properly initializing the weights for these additional channels. It is crucial to deliberate on the initialization method for these new channels when extending a model beyond the data types it was initially trained on. We investigate two strategies: setting the new weights to zero and using the mean of the original RGB weights. The rest of the model retains its baseline weights. According to Table \ref{tab:patch_init}, initializing the new weights to zero leads to improved performance across all metrics. This strategy allows the model to effectively leverage the pre-trained RGB weights, while the zero-initialized channels offer flexibility for adapting to specific tasks. Since multispectral data comprises information beyond the visible spectrum, starting with zero-initialized patch embedding weights enables the model to learn relevant characteristics from the ground up, avoiding reliance on potentially unsuitable existing features.

Furthermore, we conducted experiments to determine if all 12 multispectral bands from the dataset are necessary to achieve performance levels that exceed the baseline. Although multispectral data typically provides comprehensive information, low-quality or noisy bands can negate these advantages. As shown in Table \ref{tab:input_bands}, excluding the noisy 60\,m Coastal aerosol and Short Wave InfraRed-Cirrus bands is beneficial, and only 10 bands are needed for optimal performance. These discarded bands offer little to no beneficial input for the model's training process. Although 12 bands outperform using only RGB bands, it remains crucial to carefully choose spectral bands, as some may be noisy. By eliminating the 60\,m bands, the model focuses on the most informative spectral bands, improving overall accuracy and mean Average Precision (mAP).

\end{document}